\documentclass[twoside]{article}

\usepackage[accepted]{aistats2019}

\usepackage{amsmath}
\usepackage{amsfonts}
\usepackage{amssymb}
\usepackage{wrapfig}
\usepackage{subcaption}
\usepackage{multirow}
 \usepackage{mathtools} 

\usepackage{verbatim}

\usepackage{anyfontsize}

\usepackage{color}
\usepackage{tikz}
\usetikzlibrary{arrows,shapes,snakes,automata,backgrounds,fit,petri}
\usepackage{adjustbox}

\newcommand{\RN}[1]{%
	\textup{\lowercase\expandafter{\it \romannumeral#1}}%
}

\makeatletter
\newcommand{\distas}[1]{\mathbin{\overset{#1}{\kern\z@\sim}}}%

\usepackage{enumitem}

\usepackage{algorithm}
\usepackage{algorithmic}

\newcommand{\ie}[0]{\emph{i.e., }}

\newcommand{\eg}[0]{\emph{e.g., }}

\newcommand{\beq}{\vspace{0mm}\begin{equation}}
\newcommand{\eeq}{\vspace{0mm}\end{equation}}
\newcommand{\beqs}{\vspace{0mm}\begin{eqnarray}}
\newcommand{\eeqs}{\vspace{0mm}\end{eqnarray}}
\newcommand{\barr}{\begin{array}}
\newcommand{\earr}{\end{array}}

\newcommand{\Imat}{{\bf I}}

\newcommand{\av}[0]{{\boldsymbol{a}}}

\newcommand{\sv}[0]{{\boldsymbol{s}}}

\newcommand{\thetav}{\boldsymbol{\theta}}

\newcommand{\xiv}[0]{{\boldsymbol{\xi}}}

\newcommand{\phiv}{\boldsymbol{\phi}}
\newcommand{\psiv}{\boldsymbol{\psi}}

\newcommand{\E}{\mathbb{E}}

\newcommand{\Ncal}{\mathcal{N}}

\newcommand{\Fcal}{\mathcal{F}}

\newcommand{\Scal}{\mathcal{S}}

\newtheorem{lemma}{Lemma}

\ifx\assumption\undefined
\newtheorem{assumption}{Assumption}
\fi

\ifx\definition\undefined
\newtheorem{definition}{Definition}
\fi

\ifx\remark\undefined
\newtheorem{remark}{Remark}
\fi

\begin{document}

%
\runningtitle{Adversarial Learning of a Sampler Based on an Unnormalized Distribution}

%
\runningauthor{Chunyuan Li ~ Ke Bai ~  Jianqiao Li  ~ Guoyin Wang ~ Changyou Chen ~ Lawrence Carin}

\twocolumn[

\aistatstitle{Adversarial Learning of a Sampler\\Based on an Unnormalized Distribution}

\aistatsauthor{Chunyuan Li$^{1}$ ~ Ke Bai$^{2}$ ~  Jianqiao Li$^{2}$  ~ Guoyin Wang$^{2}$~ Changyou Chen$^{3}$ ~ Lawrence Carin$^{2}$}

\aistatsaddress{ $^{1}$Microsoft Research, Redmond \And  $^{2}$Duke University  \And $^{3}$University at Buffalo} ]

\begin{abstract}
We investigate adversarial learning in the case when only an unnormalized form of the density can be accessed, rather than samples.	
With insights so garnered, adversarial learning is extended to the case for which one has access to an unnormalized form $u(x)$ of the target density function, but no samples. Further, new concepts in GAN regularization are developed, based on learning from samples or from $u(x)$. The proposed method is compared to alternative approaches, with encouraging results demonstrated across a range of applications, including deep soft Q-learning. 
\end{abstract}

\section{Introduction}

Significant progress has been made recently on generative models capable of synthesizing highly realistic data samples \cite{gan,pixelrnn,kingma2014vae}. If $p(x)$ represents the true underlying probability distribution of data $x\in\mathcal{X}$, most of these models seek to represent draws $x\sim p(x)$ as $x=h_\theta(\epsilon)$ and $\epsilon\sim q_0$, with $q_0$ a specified distribution that may be sampled easily~\cite{gan,dcgan}. The objective is to learn $h_\theta(\epsilon)$, modeled typically via a deep neural network. Note that the model doesn't impose a form on (or attempt to explicitly model) the density function $q_\theta(x)$ used to implictly model $p(x)$.

When learning $h_\theta(\epsilon)$ it is typically assumed that one has access to a set of samples $\{x_i\}_{i=1,N}$, with each $x_i$ drawn i.i.d. from $p(x)$. While such samples are often available, there are other important settings for which one may wish to learn a generative model for $p(x)$, without access to associated samples. An important example occurs when one has access to an {\em unnormalized} distribution $u(x)$, with $p(x)=u(x)/C$ and normalizing constant $C$ unknown. The goal of sampling from $p(x)$ based on $u(x)$ is a classic problem in physics, statistics and machine learning \cite{Hastings,Gelman}. This objective has motivated theoretically exact (but expensive) methods like Markov chain Monte Carlo (MCMC)~\cite{brooks2011handbook,welling2011bayesian}, and approximate methods like variational Bayes~\cite{hoffman2013stochastic,kingma2014vae,rezende2014stochastic} and expectation propagation~\cite{minka2001expectation,li2015stochastic}. A challenge with methods of these types (in addition to computational cost/approximations) is that they are means of drawing samples or approximating density forms based on $u(x)$, but they do not directly yield a model like $x=h_\theta(\epsilon)$ and $\epsilon\sim q_0$, with the latter important for many fast machine learning implementations.

A recently developed, and elegant, means of modeling samples based on $u(x)$ is Stein variational gradient descent (SVGD)~\cite{Stein}. SVGD also learns to draw a set of samples, and an amortization step is used to learn $x=h_\theta(\epsilon)$ and $\epsilon\sim q_0$ based on the SVGD-learned samples~\cite{SteinGAN,feng2017learning,Pu_NIPS17}. Such amortization may also be used to build $h_\theta(\epsilon)$ based on MCMC-generated samples~\cite{li2017amcmc}. While effective, SVGD-based learning of this form may be limited computationally by the number of samples that may be practically modeled, limiting accuracy. Further, the two-step nature by which $x=h_\theta(\epsilon)$ is manifested may be viewed as less appealing.

In this paper we develop a new extension of generative adversarial networks (GANs)~\cite{gan} for settings in which we have access to $u(x)$, rather than samples drawn from $p(x)$. The formulation, while new, is simple, based on a recognition that many existing GAN methods constitute different means of estimating a function of a likelihood ratio~\cite{Kanamori,mohamed2016learning,uehara2016generative}. The likelihood ratio is associated with the true density function $p(x)$ and the model $q_\theta(x)$. Since we do not have access to $p(x)$ or $q_\theta(x)$, we show, by a detailed investigation of $f$-GAN~\cite{f-GAN}, that many GAN models reduce to learning $g_0(p(x)/q_\theta(x))$, where $g_0(\cdot)$ is a general monotonically increasing function. $f$-GAN is an attractive model for uncovering underlying principles associated with GANs, due to its generality, and that many existing GAN approaches may be viewed as special cases of $f$-GAN. With the understanding provided by an analysis of $f$-GAN, we demonstrate how $g_0(p(x)/q_\theta(x))$ may be estimated via $u(x)$, and an introduced reference distribution $p_r(x)$. As discussed below, the assumptions on $p_r(x)$ are that it is easily sampled, it has a known functional form, and it represents a good approximation to $q_\theta(x)$.

For the special case of variational inference for latent models, the proposed formulation recovers the adversarial variational Bayes (AVB)~\cite{AVB} setup. However, we demonstrate that the proposed approach has more applicability than inference. Specifically, we demonstrate its application to soft Q-learning~\cite{haarnoja2017reinforcement}, and it leads to the first general purpose adversarial policy algorithm in reinforcement learning.  We make a favorable comparison in this context to the aforementioned SVGD formulation.

An additional contribution of this paper concerns regularization of adversarial learning, of interest when learning based on samples or on an unnormalized distribution $u(x)$. Specifically, we develop an entropy-based regularizer. When learning based on $u(x)$, we make connections to simulated annealing regularization methods used in prior sampling-based models. We also introduce a bound on the entropy, applicable to learning based on samples or $u(x)$, and make connections to prior work on cycle consistency used in GAN regularization.

%
%
%
%
%
%
%
%
%

\begin{figure*}[t!]
	\vspace{-0mm}\centering
	\begin{tabular}{c c}		
		\hspace{-4mm}
		\includegraphics[width=9.0cm]{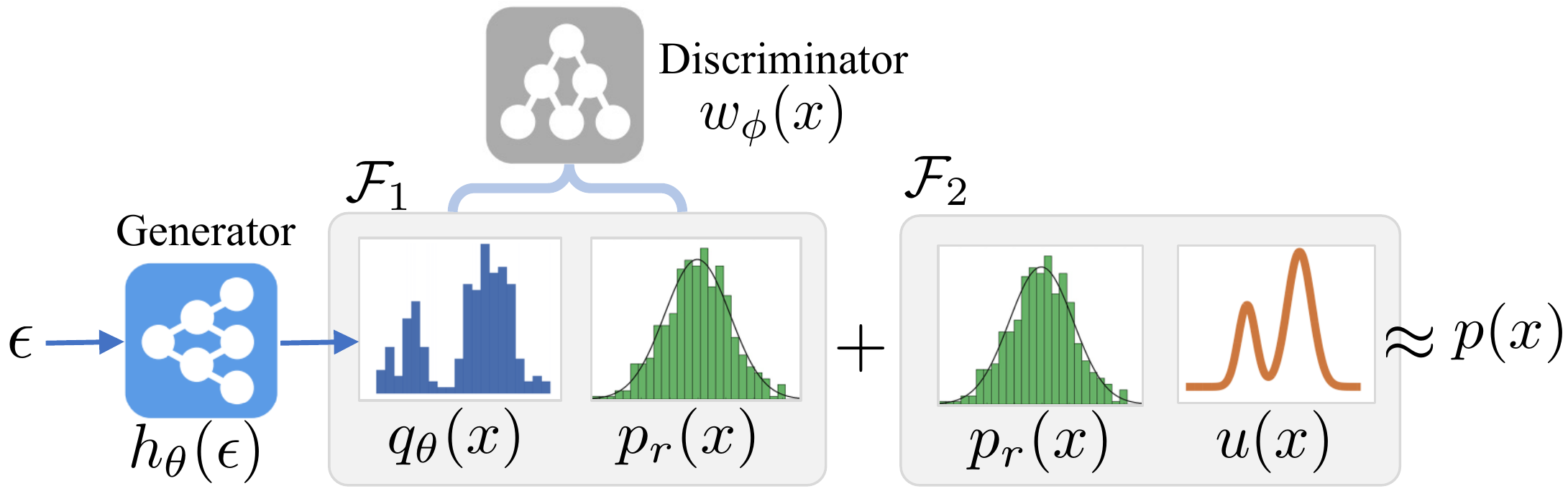} &
		\hspace{5mm}
		\includegraphics[width=5.5cm]{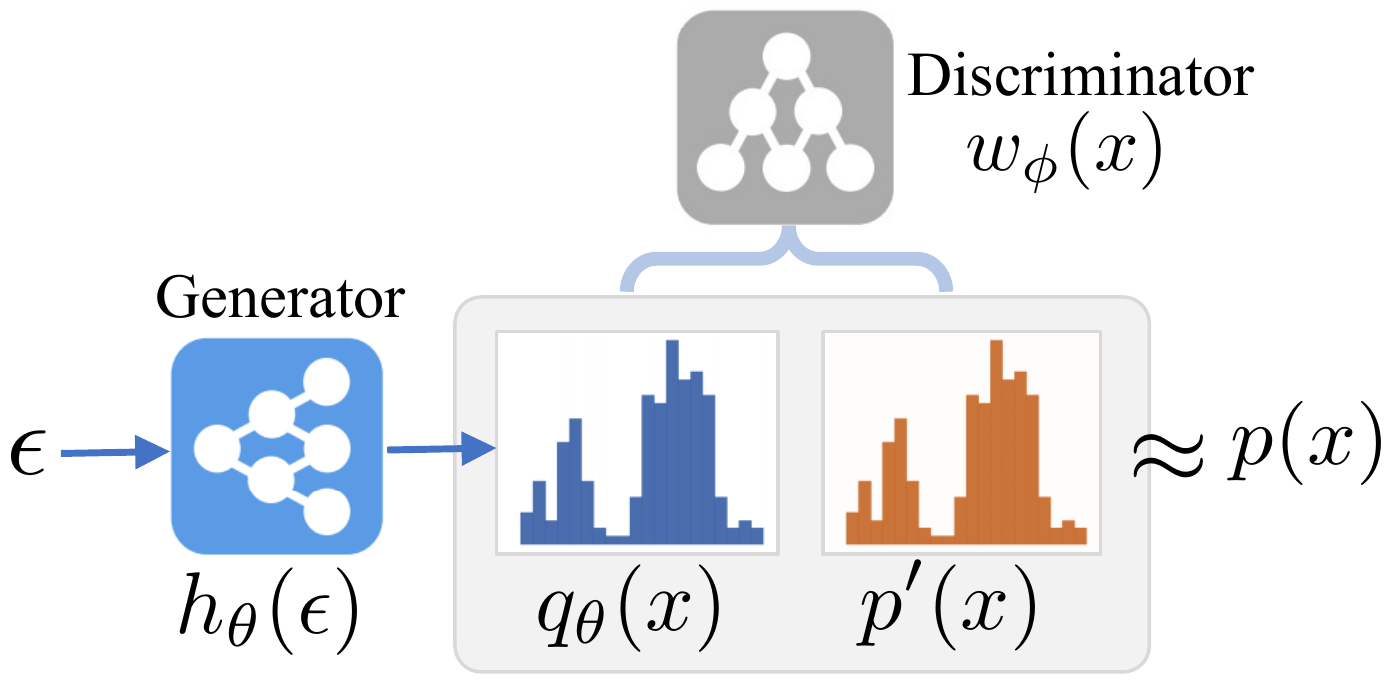} \\
		(a) Learning from an unnormalized distribution $u(x)$  \vspace{-0mm}   & 
		(b) Learning from a sample set $p'(x)$ \hspace{-0mm} 
	\end{tabular}
	\vspace{-2mm}
	\caption{Illustration of learning $q_{\theta}$ in the two different settings of the target $p(x)$. (a) Learning from an unnormalized distribution, as in RAS; (b) Learning from samples, as in the traditional GANs.}
	\vspace{-2mm}
	\label{fig:schemes}
\end{figure*}

\vspace{-0mm}
\section{Traditional GAN Learning}
\label{sec:learning_samples}
\vspace{-2mm}
We begin by discussing GAN from the perspective of the $f$-divergence \cite{likeratio}, which has resulted in $f$-GAN \cite{f-GAN}. $f$-GAN is considered because many popular GAN methods result as special cases, thereby affording the opportunity to identify generalizable components that may extended to new settings. Considering continuous probability density functions $p(x)$ and $q(x)$ for $x\in\mathcal{X}$, the $f$-divergence is defined as
$
D_f(p\|q)=\int_{\mathcal{X}} q(x) f\big[ \frac{p(x)}{q(x)}\big] dx
$, 
where $f: \mathbb{R}_+\rightarrow \mathbb{R}$ is a convex, lower-semicontinuous function satisfying $f(1)=0$. Different choices of $f[r(x)]$, with $r(x)=p(x)/q(x)$, yield many common divergences; see \cite{f-GAN} and Table \ref{tab:list}.

An important connection has been made between the $f$-divergence and generative adversarial learning, based on the inequality \cite{likeratio}
\beq
D_f(p\|q)\geq \sup_{T\in\mathcal{T}} \big[ \mathbb{E}_{x\sim p}[T(x)]-\mathbb{E}_{x\sim q}[f^*(T(x))] \big]\label{eq:bound}
\eeq
where $f^*(t)$ is the convex conjugate function, defined as $f^*(t)=\mbox{sup}_{u\in \mbox{dom}_f}\{ut-f(u)\}$, which has an analytic form for many choices of $f$ \cite{f-GAN}. Further, under mild conditions, the bound is tight when $T(x)=f^\prime \big[\frac{p(x)}{q(x)} \big]$ where $f^\prime(r)$ is the derivative of $f(r)$. Even if we know $f^\prime(r)$ we cannot evaluate $T(x)=f^\prime \big[\frac{p(x)}{q(x)} \big]$ explicitly, because $q(x)$ and/or $p(x)$ are unknown. 

Note that to compute the bound in (\ref{eq:bound}), we require expectations wrt $p$ and $q$, which we effect via sampling (this implies we only need samples from $p$ and $q$, and do not require the explicit form of the underlying distributions). Specifically, assume $p$ corresponds to the true distribution we wish to model, and $q_\theta$ is a model distribution with parameters $\theta$. We seek to learn $\theta$ by minimizing the bound of $D_f(p\|q_\theta)$ in \eqref{eq:bound}, with draws from $q_\theta$ implemented as $x=h_\theta(\epsilon)$ with $\epsilon\sim q_0$, where $q_0$ is a probability distribution that may be sampled easily (\eg uniform, or isotropic Gaussian \cite{gan}). The learning problem consists of solving
\beqs
(\hat{\theta},\hat{\phi})&=&\mbox{argmin}_\theta~\mbox{argmax}_{\phi} \big[ \mathbb{E}_{x\sim p}[T_\phi(x)]\nonumber\\
&~&~~~~~~~~~~~~~~-\mathbb{E}_{\epsilon\sim q_0}[f^*(T_\phi(h_\theta(\epsilon)))] \big]\label{eq:minimax}
\eeqs
where $T_\phi(x)$ is typically a (deep) neural network with parameters $\phi$, with $h_\theta(\epsilon)$ defined similarly. Attempting to solve (\ref{eq:minimax}) produces $f$-GAN \cite{f-GAN}.

One typically solves this minimax problem by alternating between update of $\theta$ and $\phi$ \cite{f-GAN,gan}. Note that the update of $\theta$ only involves the second term in (\ref{eq:minimax}), corresponding to $\mbox{argmax}_\theta ~\mathbb{E}_{\epsilon\sim q_0}[f^*(T_\phi(h_\theta(\epsilon)))]$. Recall that the bound in \eqref{eq:bound} is tight when $T_\phi(x)=f^\prime [p(x)/q_{\theta_{n-1}}(x)]$~\cite{nguyen2010estimating}, where $\theta_{n-1}$ represent parameters $\theta$ from the previous iteration. Hence, assuming $T_{\phi_n}(x)=f^\prime [p(x)/q_{\theta_{n-1}}(x)]$, we update $\theta$ as  
\beq
\theta_n=\mbox{argmax}_\theta~\mathbb{E}_{\epsilon\sim q_0}~g [p(h_\theta(\epsilon))/q_{\theta_{n-1}}(h_\theta(\epsilon))]\label{eq:update}
\eeq
where $g(r)=f^*(f^\prime(r))$.

Different choices of $f$ yield a different optimal function $g(r)$ (see Table \ref{tab:list}). However, in each case $\theta$ is updated such that samples from $q_\theta$ yield an increase in the likelihood ratio $r_{\theta_{n-1}}(x)=p(x)/q_{\theta_{n-1}}(x)$, implying samples from $q_\theta$ better match $p(x)$ than they do $q_{\theta_{n-1}}(x)$. Recall that the likelihood ratio $r_{\theta_{n-1}}(x)$ is the optimal means of distinguishing between samples from $p(x)$ and $q_{\theta_{n-1}}(x)$ \cite{vanTrees,neyman1933ix}. Hence, $r_{\theta_{n-1}}(x)$ is a critic, approximated through $T_{\phi_n}(x)$, that the actor $q_\theta$ seeks to maximize when estimating $\theta_n$. 

\begin{table}
\tiny
\caption{\small Functions $g(r)$ and $f(r)$ corresponding to particular $f$-GAN setups.\label{tab:list}}
\begin{tabular}{ |p{2.0cm}||p{2.95cm}||p{1.95cm}| }
 \hline
 $f$-Divergence & $f(r)$ & $g(r)$ in $\theta$ update \\
 \hline
 Kullback-Leibler (KL) & $r\log r$  & $r$ \\
 Reverse KL & $-\log r$ &   $\log r$  \\
 Squared Hellinger & $(\sqrt{r}-1)^2$ & $\sqrt{r}$ \\
 Total variation & $|r-1|/2$ & $\frac{1}{2}\mbox{sign}({r}-1)$  \\
 Pearson $\chi^2$ & $(r-1)^2$ &  $({r}-1)^2+2r$  \\
 Neyman $\chi^2$ & $(r-1)^2/r$ & $-1/r$   \\
 GAN  & $r \log r - (r + 1) \log(r + 1)$ & $-\log\big[\frac{1}{1+{r}}\big]$ \\
 \hline
\end{tabular}\vspace{-2mm}
\end{table}

We may alternatively consider
\beqs
\phi_n&=&\mbox{argmax}_\phi~\{\mathbb{E}_{x\sim p(x)} \log [\sigma (w_\phi(x))] \nonumber\\
&~&~~~+ \mathbb{E}_{\epsilon\sim q_0}\log [1-\sigma(w_\phi(h_{\theta_{n-1}}(\epsilon)))]\}\label{eq:phi}\\
\theta_n&=&\mbox{argmax}_\theta~\mathbb{E}_{\epsilon\sim q_0} g_0[w_{\phi_{n}}(h_\theta(\epsilon))]\label{eq:theta}
\eeqs
where now $g_0(r)$ is an {\em arbitrary} monotonically increasing function of $r$, $\sigma (\cdot)$ is the sigmoid function. 
From \cite{Kanamori,AVB,noise_contrastive},  the solution to (\ref{eq:phi}) is 
\vspace{-4mm}
\beqs
w_\phi(x)=\log [p(x)/q_{\theta_{n-1}}(x)], 
\eeqs
where model $w_\phi(x)$ is assumed to have sufficient capacity to represent the likelihood ratio for all $x\in\mathcal{X}$.
Hence, here $w_\phi(x)$ replaces $T_\phi(x)$ from $f$-GAN, and the solution to $w_\phi(x)$ is a particular function of the likelihood ratio. If $g_0(w_\phi(x))=w_\phi(x)$ this corresponds to learning based on minimizing the reverse KL divergence $\mbox{KL}(q_\theta\|p)$. When $g_0(\cdot)=\log [ \sigma (\cdot)]$, one recovers the original GAN \cite{gan}, for which learning corresponds to $(\hat{\theta},\hat{\phi})=\mbox{argmin}_{\theta}~\mbox{argmax}_\phi ~\{\mathbb{E}_{x\sim p(x)} \log [\sigma (w_\phi(x))] + \mathbb{E}_{\epsilon\sim q_0}\log [1-\sigma(w_\phi(h_{\theta}(\epsilon)))]\}$.

In (\ref{eq:phi})-(\ref{eq:theta}) and in $f$-GAN, respective estimation of $w_\phi(x)$ and $T_\phi(x)$ yields approximation of a function of a likelihood ratio; such an estimation appears to be at the heart of many GAN models. This understanding is our launching point for extending the range of applications of adversarial learning.

\section{Unnormalized-Distribution GAN}
\label{sec:learning_distribution}
\vspace{-3mm}
%


In the above discussion, and in virtually all prior GAN research, access is assumed to samples from target distribution $p(x)$. In many applications samples from $p(x)$ are unavailable, but the {\em unnormalized} $u(x)$ is known, with $p(x)=u(x)/C$ but with constant $C=\int u(x) {d} x$ intractable. A contribution of this paper is a recasting of GAN to cases for which we have $u(x)$ but no samples from $p(x)$, recognizing that most GAN models require an accurate estimate of the underlying likelihood ratio.

We consider the formulation in (\ref{eq:phi})-(\ref{eq:theta}) and for simplicity set $g_0(w_\phi(x))=w_\phi(x)$, although any choice of $g_0(\cdot)$ may be considered as long as its monotonically increasing. The update of $\theta$ remains as in (\ref{eq:theta}), and we seek to estimate $\log[p(x)/q_{\theta_{n-1}}(x)]$ based on knowledge of $u(x)$. Since $\log[p(x)/q_{\theta_{n-1}}(x)]=\log[u(x)/q_{\theta_{n-1}}(x)]-\log C$, for the critic it is sufficient to estimate $\log[u(x)/q_{\theta_{n-1}}(x)]$. Toward that end, 
%
we introduce a {\em reference} distribution $p_{r}(x)$, that 
$(\RN{1})$ may be sampled easily, and 
$(\RN{2})$ has an explicit functional form that may be evaluated.
The reference distribution can be connected to both importance sampling and the reference ratio method developed in bioinformatics~\cite{hamelryck2010potentials}.
We have
\vspace{-1mm}
\beq\label{eq:ref_general}
\log[\frac{u(x)}{q_{\theta_{n-1}}(x)} ]=
\underbrace{ \log  \big[\frac{ p_{r}(x) }{ q_{\theta_{n-1}} (x)  }  \big]}_{\Fcal_1} 
+ 
\underbrace{ \log  \big[\frac{ u (x) }{p_{r}(x)  }\big] }_{\Fcal_2}  
\vspace{-1mm}
\eeq
where $\Fcal_2$ may be evaluated explicitly. We learn $\Fcal_1$ via (\ref{eq:phi}), with $\mathbb{E}_{x\sim p(x)}$ changed to $\mathbb{E}_{x\sim p_r(x)}$. Therefore, learning becomes alternating between the following two updates:
\beqs
\hspace{-5mm} \phi_n &\hspace{-5mm} =& \hspace{-5mm} 
\mbox{argmax}_\phi~\{\mathbb{E}_{x\sim p_r(x)} \log [\sigma (w_\phi(x))] \nonumber\\
&~&~~~+ \mathbb{E}_{\epsilon\sim q_0}\log [1-\sigma(w_\phi(h_{\theta_{n-1}}(\epsilon)))]\}\label{eq:phi_ras}\\
\hspace{-5mm} \theta_n  &\hspace{-5mm} = & \hspace{-5mm} \mbox{argmax}_\theta~\mathbb{E}_{\epsilon\sim q_0} \Big[ w_{\phi_{n}}( h_\theta(\epsilon) ) +  \log  [\frac{ u (h_\theta(\epsilon)) }{p_{r}(h_\theta(\epsilon))  } ] \Big]  \label{eq:theta_ras}
\vspace{-2mm}
\eeqs

We call this procedure {\it reference-based adversarial sampling} (RAS) for unnormalized distributions. One should carefully note its distinction from the traditional GANs\footnote{We refer to generative models learned via samples as GAN, and generative models learned via an unnormalized distribution as RAS.}, which usually learn to draw samples to mimic the given samples of a target distribution. To illustrate the difference, we visualize the learning schemes for the two settings in Figure~\ref{fig:schemes}. 


The parameters of reference distribution $p_r(x)$ are estimated using samples from $q_{\theta}$. We consider different forms of $p_r$ depending on the application.

\hspace{-4mm}
\begin{minipage}{1.05\linewidth}
\begin{itemize}	
	\item {\bf Unconstrained domains}~
	For the case when the support $\mathcal{X}$ of the target distribution is unconstrained, 
	we model $p_r(x)$ as a Gaussian distribution with diagonal covariance matrix, with mean and variance components estimated via samples from $q_{\theta_{n-1}}$, drawn as $x=h_{\theta_{n-1}}(\epsilon)$ with $\epsilon\sim q_0$. 
	\item {\bf Constrained domains}~
	In some real-world applications the support $\mathcal{X}$ is bounded. For example, in reinforcement learning, the action often resides within a finite interval $[c_1, c_2]$. In this case, we propose to represent each dimension of $p_r$ as a generalized Beta distribution $\mbox{Beta}(\hat{\alpha}_0, \hat{\beta}_0, c_1, c_2)$. The shape parameters are estimated using method of moments: $\hat{\alpha}_0 = \bar{a} \left( \frac{ \bar{a}(1-\bar{a})}{ \bar{v}  } -1 \right)$ and $\hat{\beta}_0 = (1-\bar{a}) \left( \frac{ \bar{a}(1-\bar{a})}{ \bar{v}  } -1 \right)$, where $\bar{a} = \frac{\bar{a}^{\prime} -  c_1 }{c_2 - c_1}$ and $ \bar{v} = \frac{\bar{v}^{\prime} }{ (c_2 - c_1)^2 }$, and $\bar{a}^{\prime}$ and  $\bar{v}^{\prime}$ are sample mean and variance, respectively.
\end{itemize}
\end{minipage}


\vspace{-2mm}
\section{Entropy Regularization}
\vspace{-4mm}

Whether we perform adversarial learning based on samples from $p(x)$, as in Sec. \ref{sec:learning_samples}, or based upon an unnormalized distribution $u(x)$, as in Sec. \ref{sec:learning_distribution}, the update of parameters $\theta$ is of the form $\theta_n=\mbox{argmax}_\theta~\mathbb{E}_{\epsilon\sim q_0}g_0[\log(p(h_\theta(\epsilon))/q_{\theta_{n-1}}(h_\theta(\epsilon)))]$, where $\log(p(x)/q_{\theta_{n-1}}(x))$ is approximated as in (\ref{eq:phi}) or its modified form (for learning from an unnormalized distribution).

A well-known failure mode of GAN is the tendency of the generative model, $x=h_\theta(\epsilon)$ with $\epsilon\sim q_0$, to under-represent the full diversity of data that may be drawn $x\sim p(x)$. Considering $\theta_n=\mbox{argmax}_\theta~\mathbb{E}_{\epsilon\sim q_0}g_0[\log(p(h_\theta(\epsilon))/q_{\theta_{n-1}}(h_\theta(\epsilon)))]$, $\theta_n$ will seek to favor synthesis of data $x$ for which $q_{\theta_{n-1}}(x)$ is small and $p(x)$ large. When learning $q_\theta(x)$ in this manner, at iteration $n$ the model $q_{\theta_{n}}$ tends to favor synthesis of a subset of data $x$ that are probable from $p(x)$ and less probable from $q_{\theta_{n-1}}(x)$. This subset of data that $q_{\theta_{n}}$ models well can change with $n$, with the iterative learning continuously moving to model a subset of the data $x$ that are probable via $p(x)$. This subset can be very small, in the worst case yielding a model that always generates the {\em same} single data sample that looks like a real draw from $p(x)$; in this case $h_\theta(\epsilon)$ yields the same or near-same output for all $\epsilon\sim q_0$, albeit a realistic-looking sample $x$. 

To mitigate this failure mode, it is desirable to add a regularization term to the update of $\theta$, encouraging that the entropy of $q_{\theta_{n}}$ be large at each iteration $n$, discouraging the model from representing (while iteratively training) a varying small subset of the data supported by $p(x)$. Specifically, consider the regularized update of \eqref{eq:theta} as:
\beq
\theta_n=\mbox{argmax}_\theta~\mathbb{E}_{\epsilon\sim q_0}g_0[w_{\phi_n}(h_\theta(\epsilon))]+\beta H(q_\theta) \label{eq:reg}
\eeq
where $H(q_\theta)$ represents the entropy of the distribution $q_\theta$, for $\beta>0$. The significant challenge is that $H(q_\theta)=-\mathbb{E}_{x\sim q_\theta}\log q_\theta(x)$, but by construction we lack an explicit form for $q_\theta(x)$, and hence the entropy may not be computed directly. Below we consider two means by which we may approximate $H$, one of which is explicitly appropriate for the case in which we learn based upon the unnormalized $u(x)$, and the other of which is applicable to whether we learn via samples from $p(x)$ or based on $u(x)$. 

In the case for which $p(x)=u(x)/C$ and $u(x)$ is known, we may consider approximating or replacing $H(q_\theta)$ with $-\mathbb{E}_{x\sim q_\theta}\log u(x)+\log C$, and the term $\log C$ may be ignored, because it doesn't impact the regularization in (\ref{eq:reg}); we therefore replace the entropy $H(q_\theta)$ with the cross entropy $-\mathbb{E}_{x\sim q_\theta}\log p(x)$. The first term in (\ref{eq:reg}) tends to encourage the model to learn to draw samples where $p(x)$, or $u(x)$, is large, while the second term discourages over-concentration on such high-probablity regions, as $-\mathbb{E}_{x\sim q_\theta}\log p(x)$ becomes large when $q_\theta$ encourages samples near lower probability regions of $p(x)$. This will ideally yield a spreading-out of the samples encouraged by $q_\theta$, with high-probability regions of $p(x)$ modeled well, but also regions spreading out from these high-probability regions.

To gain further insight into (\ref{eq:reg}), we again consider the useful case of $g_0[w_{\phi_n}(h_\theta(\epsilon))]=w_{\phi_n}(h_\theta(\epsilon))$ and assume the ideal solution $w_{\phi_n}(x)=\log [p(x)/q_{\theta_{n-1}}(x)]$. In this case cross-entropy-based regularization may be seen as seeking to maximize wrt $\theta$ the function
\beqs
\mathbb{E}_{x\sim q_\theta}\log[p(x)/q_\theta(x)]&-&\beta \mathbb{E}_{x\sim q_\theta}\log p(x)\nonumber\\&=&\mathbb{E}_{x\sim q_\theta}\log[p(x)^{1-\beta}/q_\theta(x)]\nonumber
\eeqs
For the special case of $p(x)=\exp[-E(x)]/C$, with $E(x)>0$ an ``energy'' function, we have $p(x)^{1-\beta}=\exp[-\frac{1}{T_\beta}E(x)]/C$ with $T_\beta=1/(1-\beta)$. Hence, the cross-entropy regularization is analogous to annealing, with $\beta\in[0,1)$; $\beta\rightarrow 1_-$ corresponds to high ``temperature'' $T_\beta$, which as $\beta\rightarrow 0_+$ is lowered and with $p(x)^{1-\beta}\rightarrow p(x)$. When $\beta>0$ the peaks in $p(x)$ are ``flattened out,'' allowing the model to yield samples that ``spread out'' and explore the diversity of $p(x)$. This interpretation suggests learning via (\ref{eq:reg}), with the cross-entropy replacement for $H(q_\theta)$, with $\beta$ near 1 one at the start, and progressively reducing $\beta$ toward $0$ (corresponding to lowering temperature $T_\beta$).

The above setup assumes we have access to $u(x)$, which is not the case when we seek to learn $q_\theta$ based on samples of $p$. Further, rather than replacing $H(q_\theta)$ by the cross-entropy, we may wish to approximate $H(q_\theta)$ based on samples of $q_\theta$, which we have via $x=h_\theta(\epsilon)$ with $\epsilon\sim q_0$ (with this estimated via samples from $p(x)$ or based on $u(x)$). Toward that end, consider the following lemma.

\begin{lemma} \label{lem:bound}Let $t_{\xi}(\epsilon|x)$ be a probabilistic inverse mapping associated with the generator $q_{\theta} (x)$, with parameters $\xi$. The mutual information between $x$ and $\epsilon$ satisfies
	\beq ~\label{eq:op_entropy}
	\mbox{I}(x;\epsilon)=\mbox{H}(q_{\theta})\geq \mbox{H}(q_0)+\mathbb{E}_{\epsilon\sim q_0}\log t_{\xi}(\epsilon|h_{\theta}(\epsilon)).
	\eeq
\end{lemma}\vspace{-2mm}
The proof is provided in the Supplement Material (SM). 
Since ${H}(q_0)$ is a constant wrt $(\theta,\xi)$, one may seek to maximize $\mathbb{E}_{\epsilon\sim q_0}\log t_{\xi} (\epsilon| h_{\theta}(\epsilon))$ to increase the entropy ${H}(q_\theta)$. Hence, in (\ref{eq:reg}) we replace the entropy term with $\mathbb{E}_{\epsilon\sim q_0}\log t_{\xi} (\epsilon| h_{\theta}(\epsilon))$.
%

In practice we consider $t_{\xi}(\epsilon|x)=\mathcal{N}(\epsilon;\mu_{\xi}(x), I)$, where here $I$ is the identity matrix, and $\mu_{\xi}(x)$ is a vector mean. Hence, $H(q_\theta)$ in (\ref{eq:reg}) is replaced by $-\mathbb{E}_{\epsilon\sim q_0}\|\epsilon-\mu_{\xi}(h_\theta(\epsilon))\|_2^2$. Note that a failure mode of GAN, as discussed above, corresponds to many or all $\epsilon\sim q_0$ being mapped via $x=h_\theta(\epsilon)$ to the same output. This is discouraged via this regularization, as such behavior makes it difficult to simultaneously minimize $\mathbb{E}_{\epsilon\sim q_0}\|\epsilon-\mu_{\xi}(h_\theta(\epsilon))\|_2^2$.
%
This regularization is related to cycle-consistency~\cite{ALICE}. 
However, the justification of the negative cycle-consistency as a lower bound of ${H}(q_\theta)$ is deemed a contribution of this paper (not addressed in \cite{ALICE}). 

\vspace{-2mm}
\section{Related Work}
\vspace{-2mm}

\textbf{Use of a reference distribution} We have utilized a readily-sampled reference distribution, with known density function $p_r(x)$, when learning to sample from an unnormalized distribution $u(x)$. The authors of \cite{noise_contrastive} also use such a reference distribution to estimate the probability distribution associated with observed data samples. However, \cite{noise_contrastive} considered a distinct problem, for which one wished to fit observed samples to a {\em specified} unnormalized distribution. Here we employ the reference distribution in the context of learning to sample from a known $u(x)$, with no empirical samples from $p(x)$ provided. 

\textbf{Adversarial variational Bayes} In the context of variational Bayes analysis, the adversarial variational Bayes (AVB)~\cite{AVB} was proposed for posterior inference of variational autoencoders (VAEs)~\cite{kingma2014vae}. Assume we are given a {\em parametric} generative model $p_\theta(x|z)$ with prior $p(z)$ on latent code $z$, designed to model observed data samples $\{x_i\}_{i=1,N}$. There is interest in designing an inference arm, capable of efficiently inferring a distribution on the latent code $z$ given observed data $x$. Given observed $x$, the posterior distribution on the code is $p_\theta(z|x)=p_\theta(x|z)p(z)/p_\theta(x)\propto p_\theta(x|z)p(z)$, 
where $p_\theta(x)=\int p_\theta(x|z)p(z)dz$, and $u_\theta(z;x)=p_\theta(x|z)p(z)$ represents an {\em unnormalized} distribution of the latent variable $z$, which also depends on the data $x$.

One may show that if the procedure in Sec. \ref{sec:learning_distribution} is employed to draw samples from $p_\theta(z|x)$, based on the unnormalized $u_\theta(z;x)$, one exactly recovers AVB~\cite{AVB}. The AVB considered $g_0(w_\phi)=w_\phi$ within our framework. 
We do not consider the application to inference with VAEs, as the experiments in \cite{AVB} are applicable to the framework we have developed. The generality of the RAS is made more clear in our paper.
We show its applicability to reinforcement learning in Sec. \ref{sec:experiments}, and broaden the discussion on the type of adaptive reference distributions in Sec. 3, with extensions to constrained domain sampling.

\textbf{Regularization} The term $\mathbb{E}_{\epsilon\sim q_0}\log t_\xi(\epsilon|h_\theta(\epsilon))$ employed here was considered in \cite{ALICE,sVAE,cycleGAN}, but the use of it as a bound on the entropy of $q_\theta$ is new. From Lemma \ref{lem:bound} we see that $\mathbb{E}_{\epsilon \sim q_0}\log t_\xi(\epsilon|h_\theta(\epsilon))$ is also a bound on the mutual information between $x$ and $\epsilon$, maximization of which is the same goal as InfoGAN \cite{infogan}. However, unlike in \cite{infogan}, here the mapping $\epsilon\rightarrow x$ is deterministic, where in InfoGAN it is stochastic. Additionally, the goal here is to encourage diversity in generated $x$, which helps mitigate mode collapse, where in InfoGAN the goal was to discover latent semantic concepts. 
%

\begin{figure*}[t!]
	\centering
	\begin{minipage}[t]{.6\textwidth}
		\centering
		\vspace{0pt}
		\includegraphics[width=7.5cm]{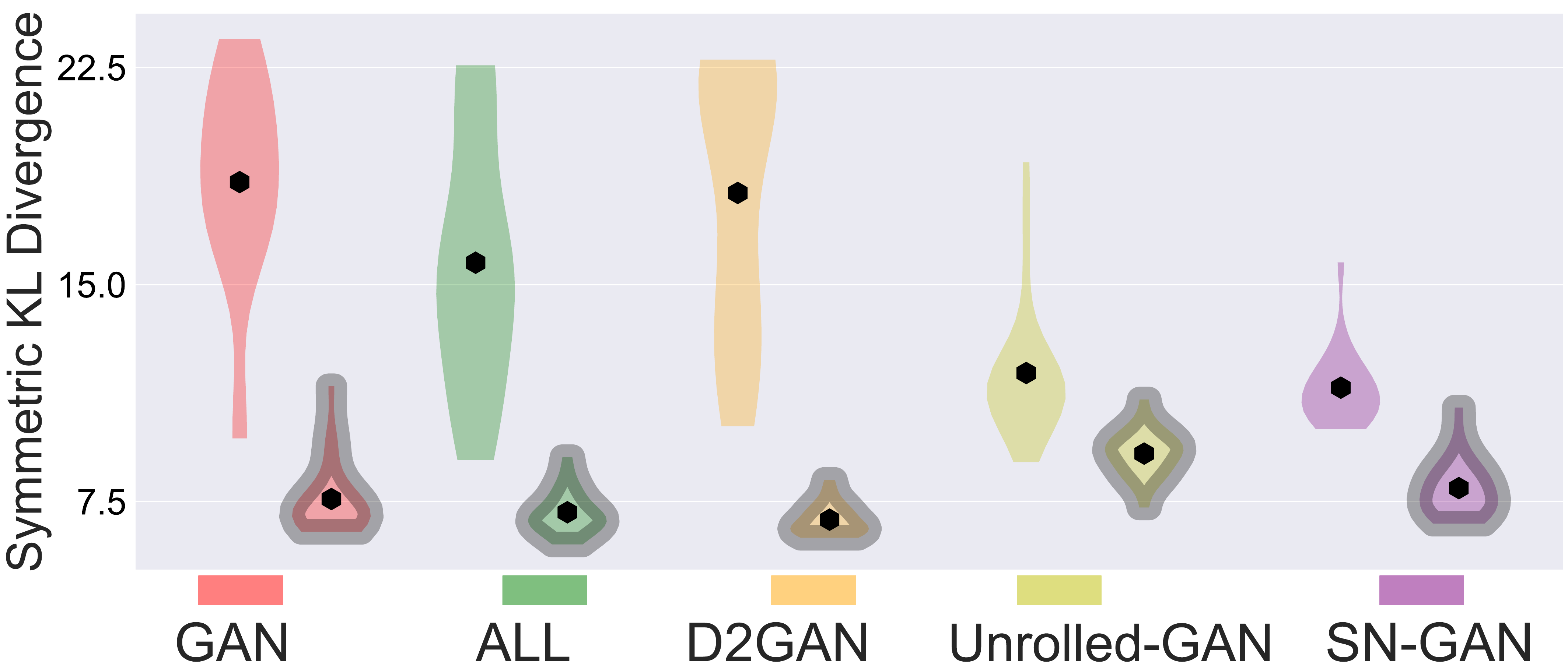}  
		\caption{Comparison of different GAN variants. The GAN models and corresponding entropy-regularized variants are visualized in the same color; in each case, the left result is unregularized, and the right employs entropy regularization. The black dots indicate the means of the distributions.}
		\label{fig:8gmm_skl}		
	\end{minipage}\hfill
	\begin{minipage}[t]{.36\textwidth}
		\centering
		\vspace{0pt}
		\begin{tabular}{c c}		
			\hspace{-0mm}
			\includegraphics[width=2.05cm]{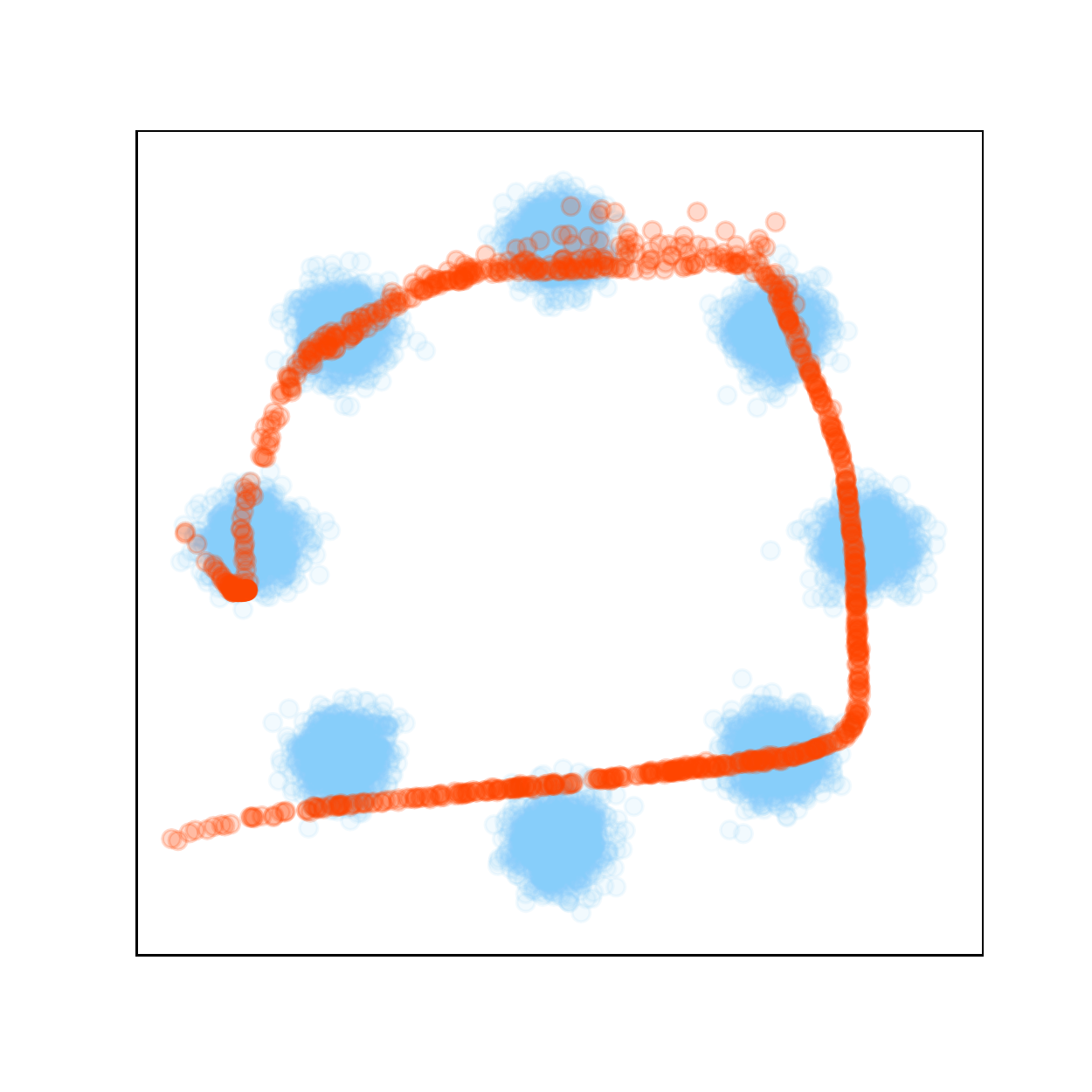}  &
			\hspace{-0mm}
			\includegraphics[width=2.05cm]{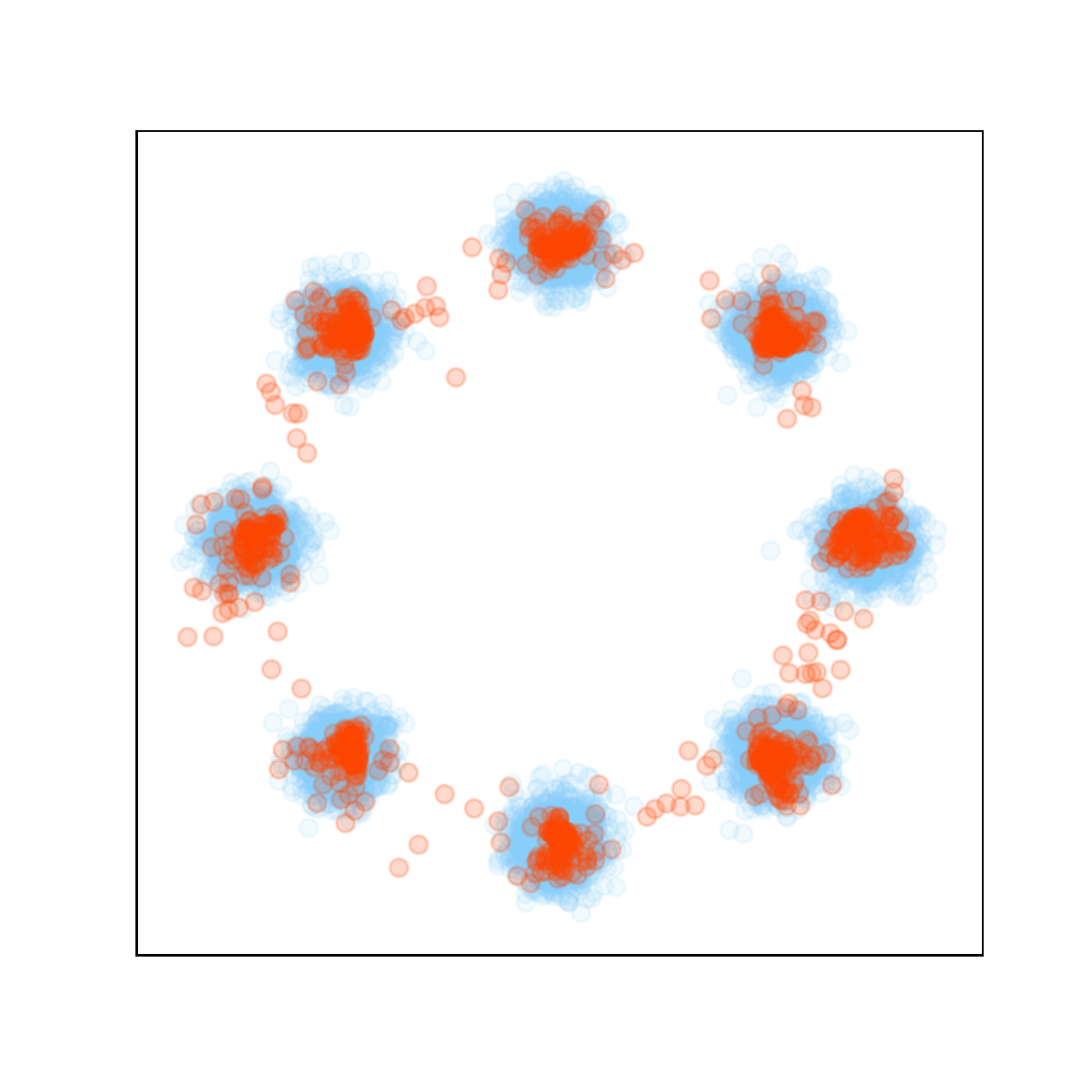} \vspace{-1mm} \\
			{\small (a) GAN \hspace{-0mm} }& \hspace{-0mm} 
			{\small 	(b) GAN-E }\hspace{-0mm}  \hspace{-0mm} 	\\
			\hspace{-0mm}
			\includegraphics[width=2.05cm]{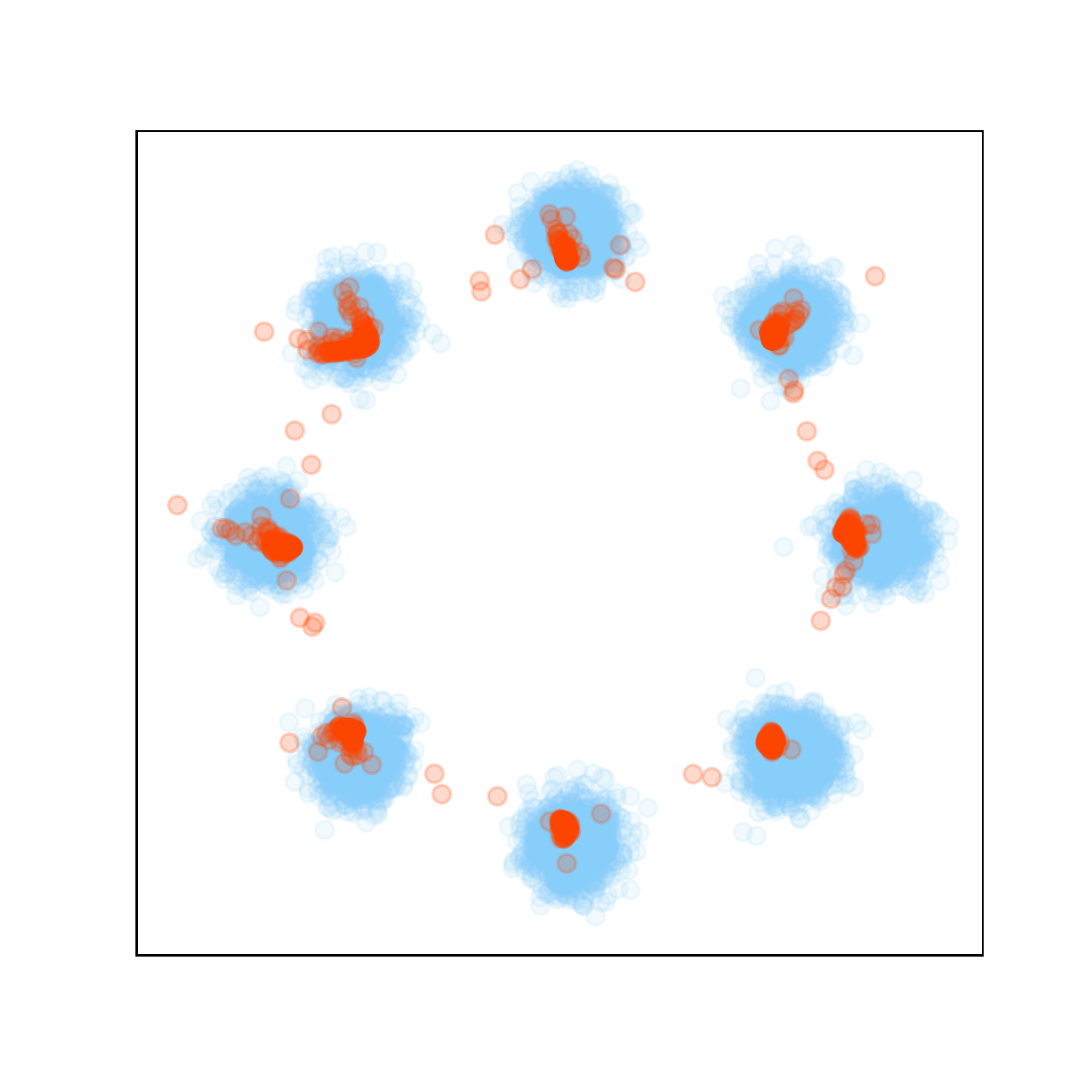}  
			&
			\hspace{-0mm}
			\includegraphics[width=2.06cm]{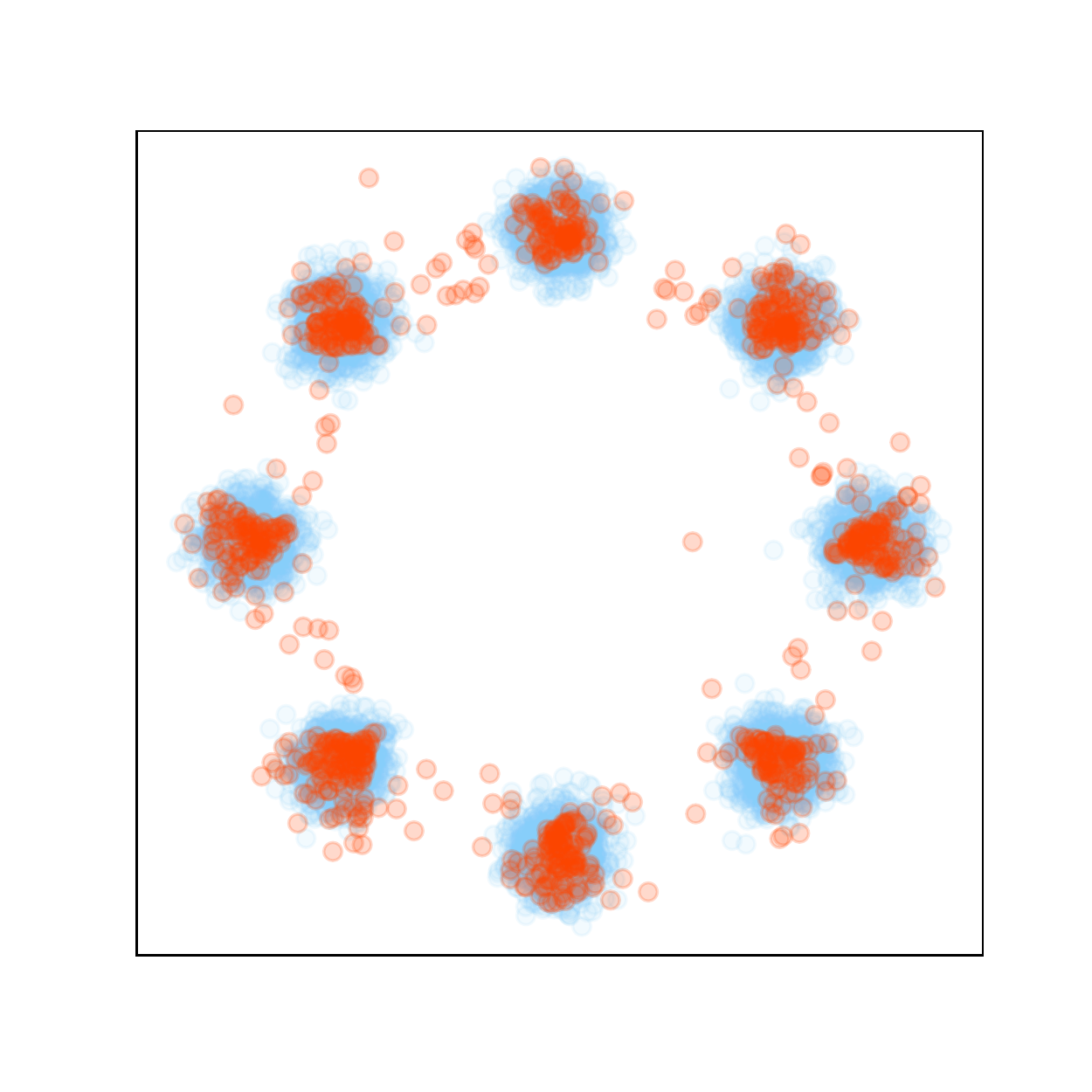}  \vspace{-1mm}
			\\
			\hspace{-0mm} 
			{\small 	(c) SN-GAN} \hspace{-0mm} & \hspace{-0mm} 
			{\small (d) SN-GAN-E}
		\end{tabular}
		\vspace{-2mm} 
		\caption{Generated samples.}
		\label{fig:8gmm_generated_samples}
	\end{minipage}
\vspace{-2mm} 
\end{figure*}

\textbf{Stein variational gradient descent (SVGD)} In the formulation of (\ref{eq:phi})-(\ref{eq:theta}), if one sets $g_0(w_\phi)=w_\phi$, then the learning objective corresponds to minimizing the reverse KL divergence $\mbox{KL}(q_\theta(x)\|p(x))$. SVGD \cite{Stein} also addresses this goal given unnormalized distribution $u(x)$, with $p(x)=u(x)/C$. Like for the proposed approach, the goal is not to explicitly learn a functional form for $q_{\theta}(x)$, rather the goal of SVGD is to learn to draw samples from it. We {\em directly} learn a sampler model via $x=h_\theta(\epsilon)$ and $\epsilon\sim q_0$, where in \cite{SteinGAN} a specified set of samples is adjusted sequentially to correspond to draws from the unnormalized distribution $u(x)$. In this setting, one assumes access to a set of samples $\{x_i\}$ drawn from some distribution, and these samples are updated deterministically as
$
x_i^\prime=x_i+\mu\gamma(x_i)
$
where $\mu>0$ is a small step size, and $\gamma(x)$ is a nonlinear function, assumed described by a reproducing kernel Hilbert space (RKHS) with given kernel $k(x,x^\prime)$. In this setting, the samples are updated $\{x_i\}\rightarrow\{x_i^\prime\}$, with a deterministic function  
$\gamma(x)$ that is evaluated in terms of $u(x)$ and $\nabla_x u(x)$. While this process is capable of transforming a specific set of samples such that they ultimately approximate samples drawn from $p(x)$, we do not have access to a model $x=h_\theta(\epsilon)$ that allows one to draw new samples quickly, on demand. Consequently, within the SVGD framework, a model $x=h_\theta(\epsilon)$ is learned separately as a second ``amortization'' step. The two-step character of SVGD should be contrasted with the direct approach of the proposed model to learn $x=h_\theta(\epsilon)$. SVGD has been demonstrated to work well, and therefore it is a natural model against which to compare, as considered below.\\

\vspace{-2mm}
\section{Experimental Results}\label{sec:experiments}
\vspace{-2mm}

%

The Tensorflow code to reproduce the experimental results 
is at github\footnote{$\mathtt{https://github.com/ChunyuanLI/RAS}$}.

\vspace{-2mm}
\subsection{Effectiveness of Entropy Regularization}
\vspace{-2mm}\label{sec:experiments_sample}
\subsubsection{Learning based on samples} 
\vspace{-0mm}
We first demonstrate that the proposed entropy regularization improves mode coverage when learning based on samples.
Following the design in~\cite{metz2017unrolled}, we consider a synthetic dataset of samples drawn from a 2D mixture of 8 Gaussians. 
The results on real datasets are reported in SM.

We consider the original GAN and three state-of-the-art GAN variants: Unrolled-GAN~\cite{metz2017unrolled}, D2GAN~\cite{D2GAN} and Spectral Normalization (SN)-GAN~\cite{miyato2018spectral}. For simplicity, we consider the case when $g_0(\cdot)$ is an identity function, and this form of GAN is denoted as {\it adversarially learned likelihood-ratio} (ALL) in Fig.~\ref{fig:8gmm_skl}. 

For all variants, we study their entropy-regularized versions, by adding the entropy bound in~\eqref{eq:op_entropy}, when training the generator. 
If not specifically mentioned, we use a {\it fix-and-decay} scheme for $\beta$ for all experiments: In total $T$ training iterations, we first fix $\beta=1$ in the first $T_0$ iteration, then linearly decay it to 0 in the rest $T-T_0$ iterations. On this 8-Gaussian dataset, $T=50$k and $T_0=10$k.

Twenty runs were conducted for each algorithm. Since we know the true distribution in this case, we employ the symmetric KL divergence as a metric to quantitatively compare the quality of generated data.  In Fig.~\ref{fig:8gmm_skl} we report the distribution of divergence values for all runs.
We add the entropy bound to each variant, and visualize their results as violin plots with gray edges (the color for each variant remains for comparison). 
The largely decreased mean and reduced variance of the divergence show that the entropy annealing yields significantly more consistent and reliable solutions, across all methods.
%
%
We plot the generated samples in Fig.~\ref{fig:8gmm_generated_samples}.  We visualize the generated samples of the original GAN in Fig.~\ref{fig:8gmm_generated_samples}(a). The samples ``struggle'' between covering all modes and separating modes. This issue is significantly reduced by ALL with entropy regularization, as shown in Fig.~\ref{fig:8gmm_generated_samples}(b). 
SN-GAN (Fig.~\ref{fig:8gmm_generated_samples}(c)) generates samples that concentrate only around the centroid of the mode. However, after adding our entropy regularizer (Fig.~\ref{fig:8gmm_generated_samples}(d)), the issue is alleviated and the samples spread out.

\vspace{-0mm}
\subsubsection{Learning based on an unnormalized distribution} 
\begin{figure}[t!]
	\vspace{-0mm}
	\begin{tabular}{c c c c}		
		\hspace{-4mm}
		\includegraphics[width=1.96cm]{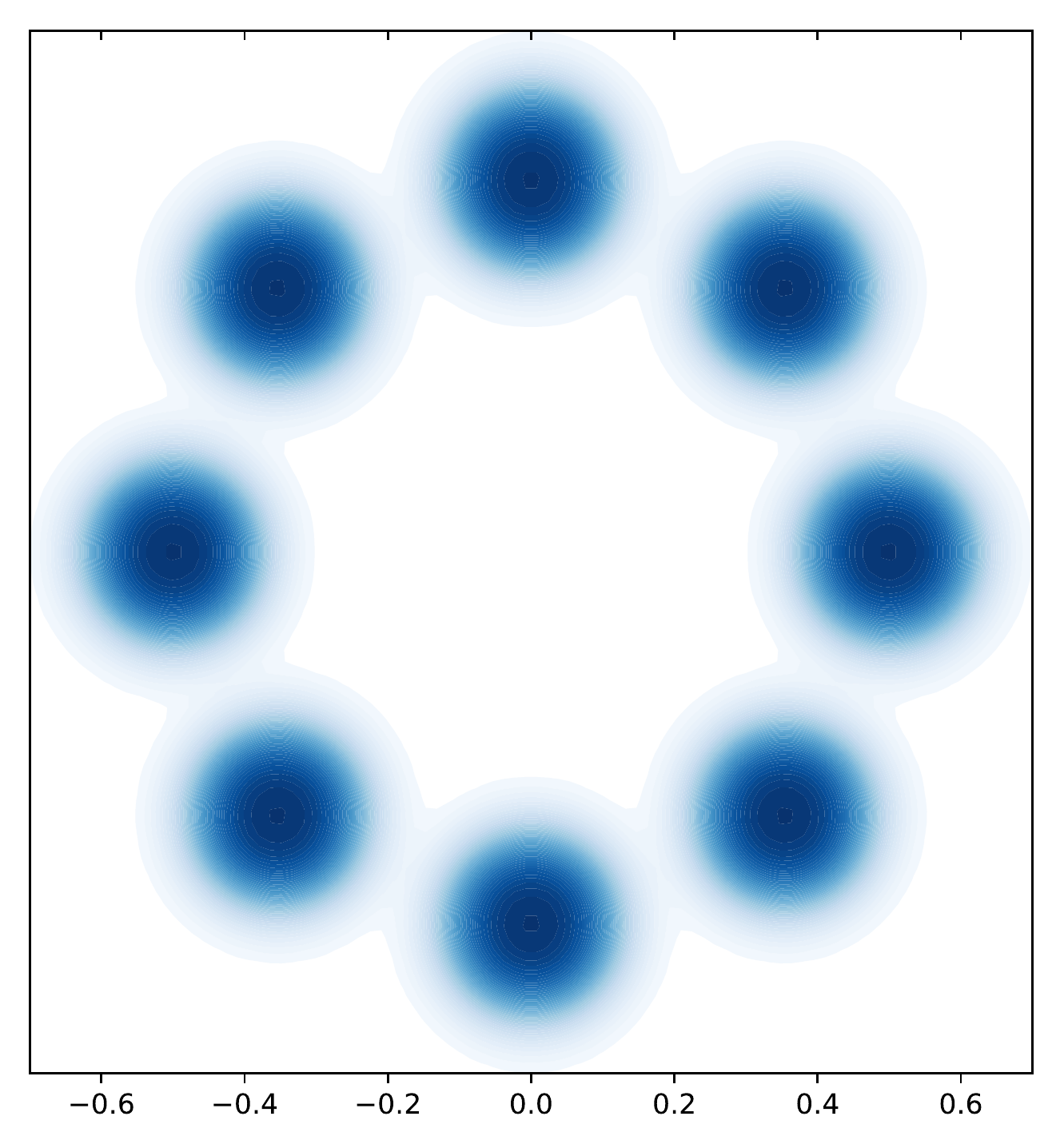}  &
		\hspace{-5mm}
		\includegraphics[width=2.0cm]{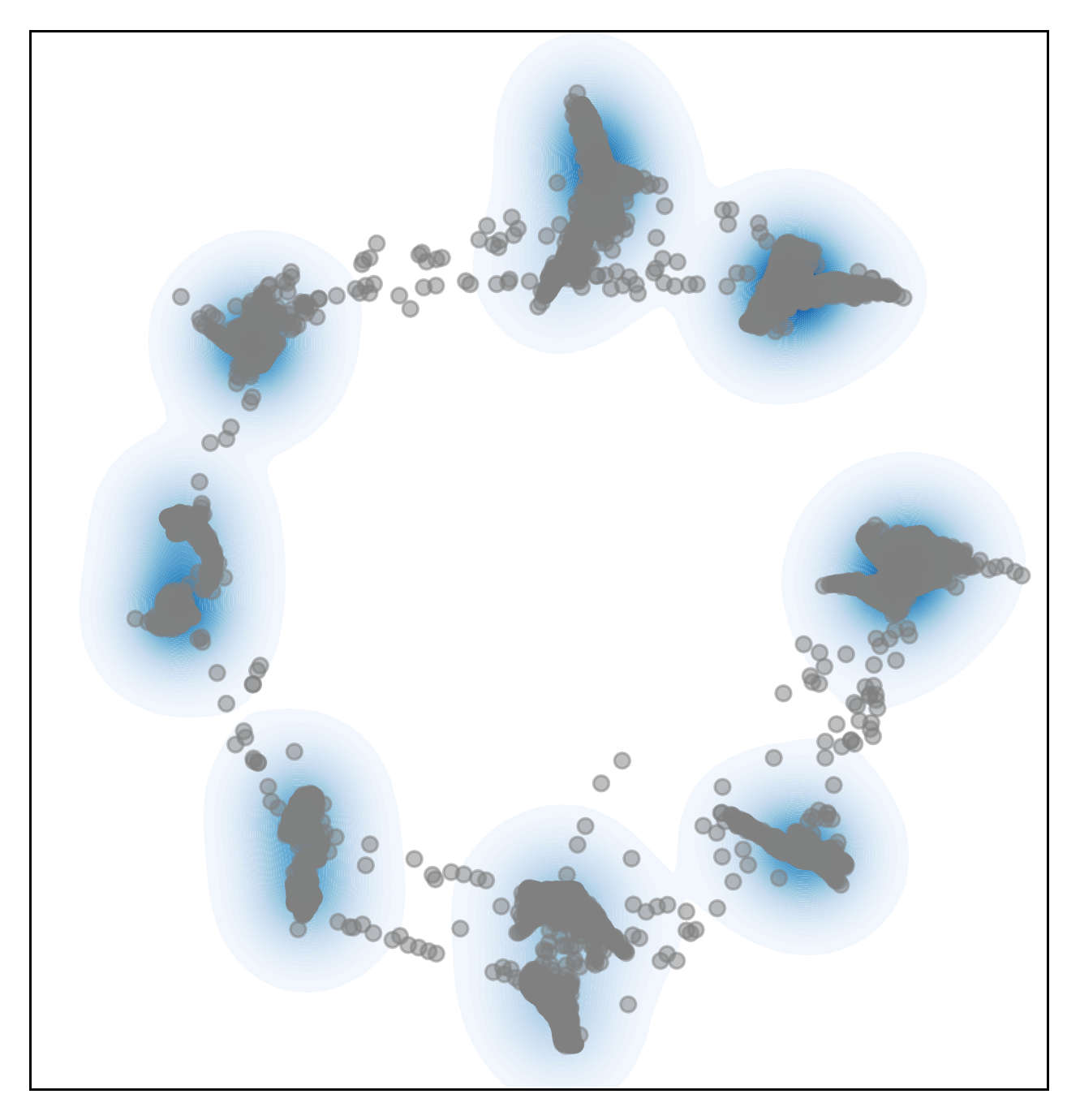}  &
		\hspace{-6mm}
		\includegraphics[width=2.0cm]{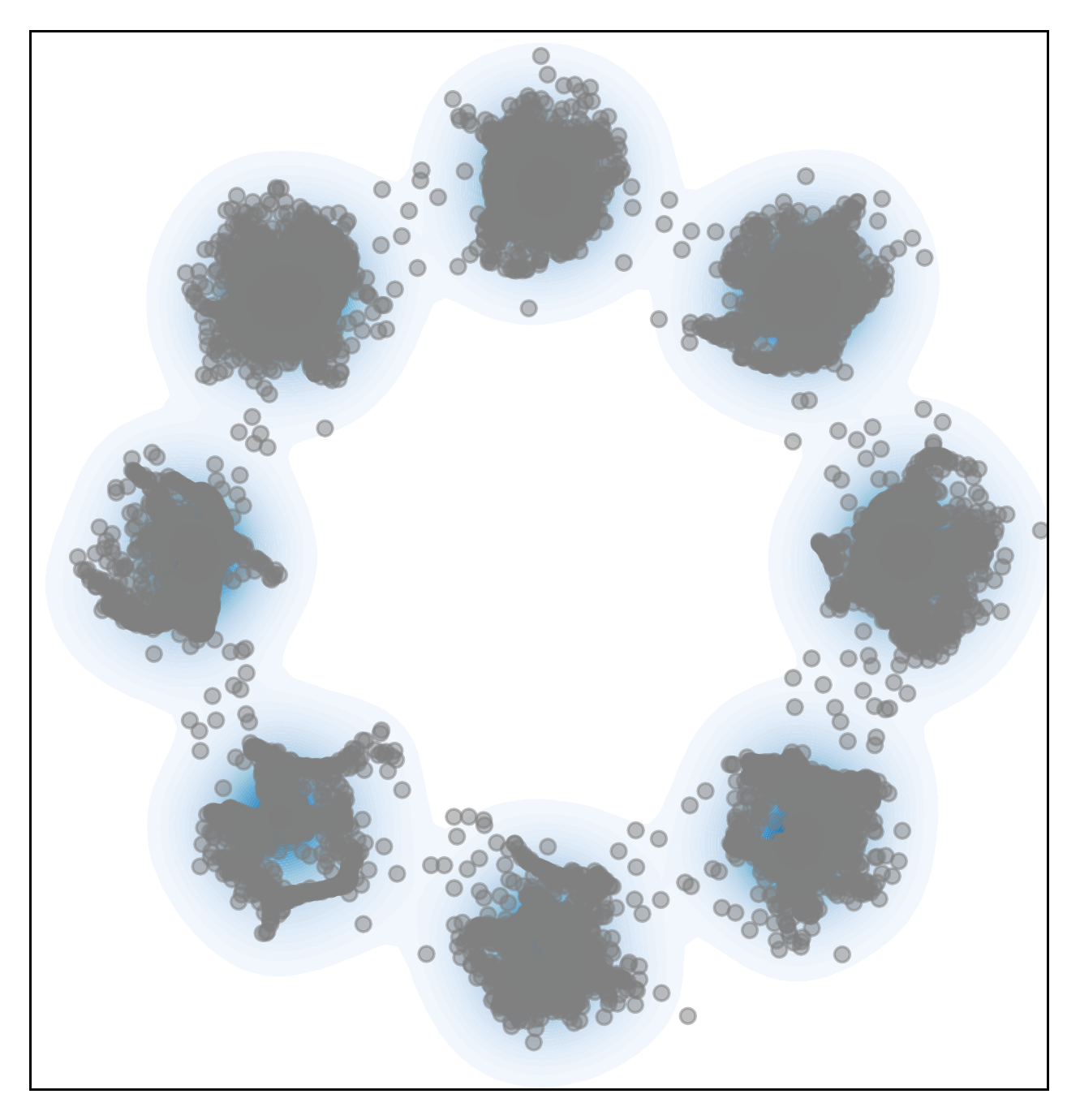}  &
		\hspace{-7mm}
		\includegraphics[width=2.0cm]{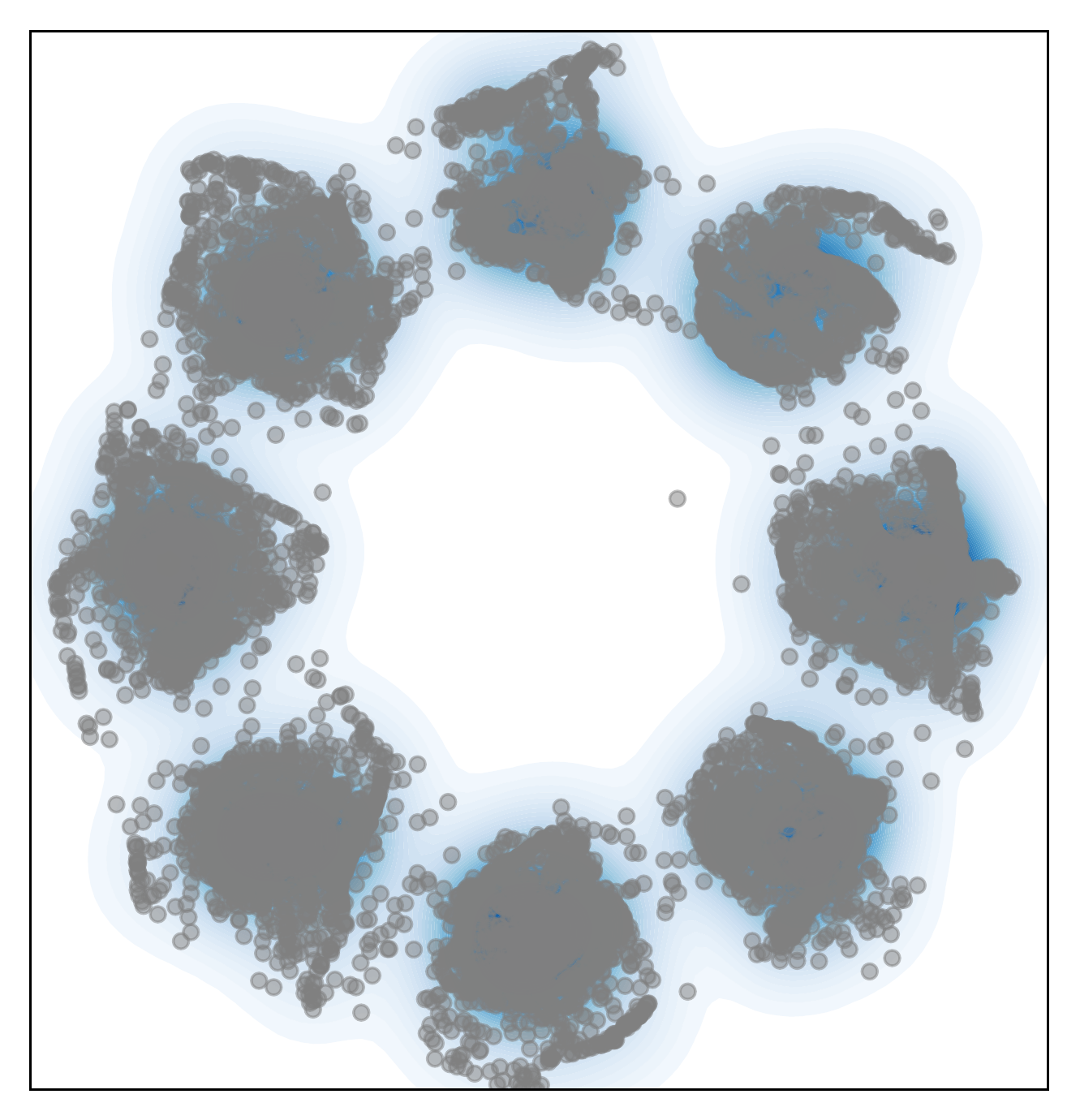} \vspace{-0mm} \\	
		(a) Target \vspace{-0mm} & \hspace{-6mm}
		(b) RAS\hspace{-0mm} 	&	 \hspace{-6mm}	
		(c) RAS+E$_{cc}$ \vspace{-0mm} & \hspace{-6mm}
		(d) RAS+E$_{ce}$ \hspace{-0mm} \hspace{-2mm} 
	\end{tabular}
	\vspace{-4mm}
	\caption{Entropy regularization for unnormalized distributions.}
	\vspace{-2mm}
	\label{fig:entropy}
\end{figure}

\vspace{-0mm}
When the unnormalized form of a target distribution is given, we consider two types of entropy regularization to improve our RAS algorithm: 
$(\RN{1})$ E$_{cc}$: the cycle-consistency-based regularization;
$(\RN{2})$ E$_{ce}$: the cross-entropy-based regularization. To clearly see the advantage of the regularizers, we fix $\beta=0.5$ in this experiment. Figure~\ref{fig:entropy} shows the results, with each case shown in one row. 
The target distributions are shown in column (a),  the sampling results of RAS are shown in column (b). RAS can reflect the general shape of the underlying distribution, but tends to concentrate on the high density regions. The two entropy regularizers are shown in (c) and (d). The entropy encourages the samples to spread out, leading to better approximation, and E$_{cc}$ appears to yield best performance. 



\vspace{-0mm}
\subsubsection{Comparison of two learning settings}
\vspace{-2mm}
In traditional GAN learning, we have a finite set of $N$ samples with the {\em empirical} distribution $p^{\prime}(x)$ to learn from, each sample drawn from the true distribution $p(x)$. 
It is known that the optimum of GANs yields the marginal distribution matching $q_{\theta}(x)=p^{\prime}(x)$~\cite{gan}; it also implies that the performance of $q_{\theta}(x)$ in is limited by $p^{\prime}(x)$. 
In contrast, when we learn from an unnormalized form as in RAS, the likelihood ratio is estimated using samples drawn from $p_r(x)$ and from $q_\theta$. Hence, we can draw as many samples as desired to get an accurate likelihood-ratio estimation, which further enables $q_\theta$ to approach $p(x)$. This means RAS can potentially provide better approximation, when $u(x)$ is available.


%
\begin{wrapfigure}{R!}{4.1cm}
	\vspace{-5mm}
	\hspace{-4mm}
	\includegraphics[width=4.25cm]{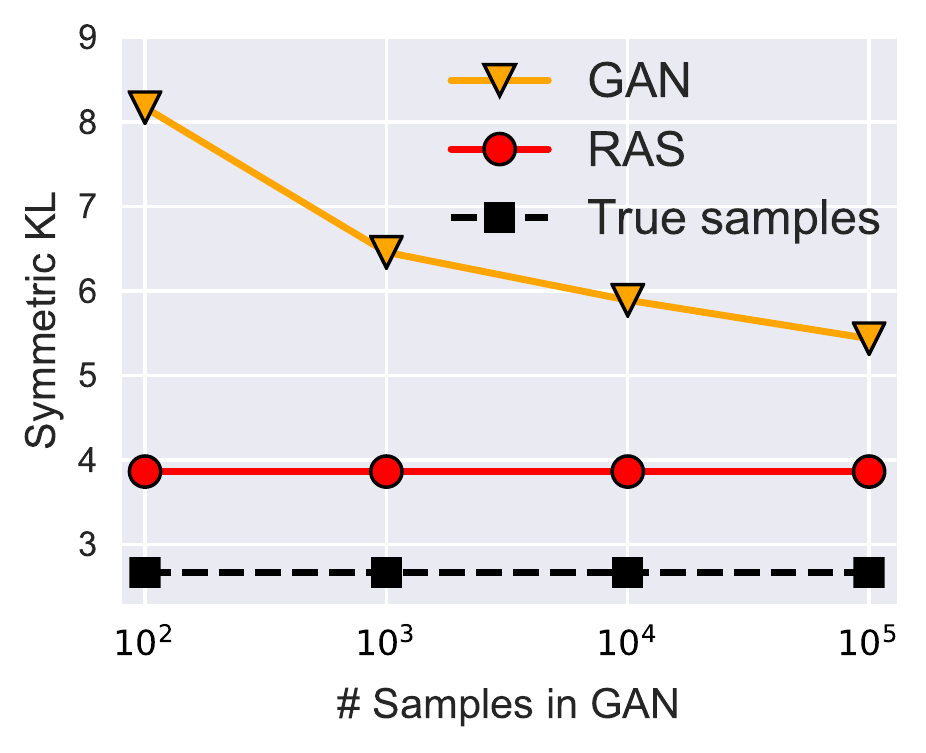}  
	\vspace{-3mm}
	\caption{\small Comparison of learning via GAN and RAS.}
	\vspace{-4mm}
	\label{fig:gan_ras}
\end{wrapfigure}

We demonstrate this advantage on the above 8-Gaussian distribution. We train GAN on $p^{\prime}(x)$ with $N=100$, $1000, 10000, 100000$ samples, and train RAS on $u(x)$. Note that the samples from $p_r$ and $q_\theta$ are drawn in an online fashion to train RAS. With an appropriate number of iterations ($T\!=\!50$k) to assure convergence, in total $T\! \cdot \! B\approx50M$ samples were used to estimate the likelihood ratio in~\eqref{eq:phi_ras}, where $B\!\!=\!\!1024$ is the minibatch size.

In the evaluation stage, we draw 20k samples from $q_{\theta}$ for each model, and compute the symmetric KL divergence against the true distribution. The results are shown in Figure~\ref{fig:gan_ras}. 
As an illustration for the ideal performance, we draw 20k samples from the target distribution and show its divergence as the black line.
The GAN gradually performs better, as more target samples are available in training. However, they are still worse than RAS by a fairly large margin. 


\begin{figure*}[t!]
	\vspace{-0mm}
	\begin{tabular}{c c c c}		
		\hspace{-4mm}
		\includegraphics[width=4.2cm]{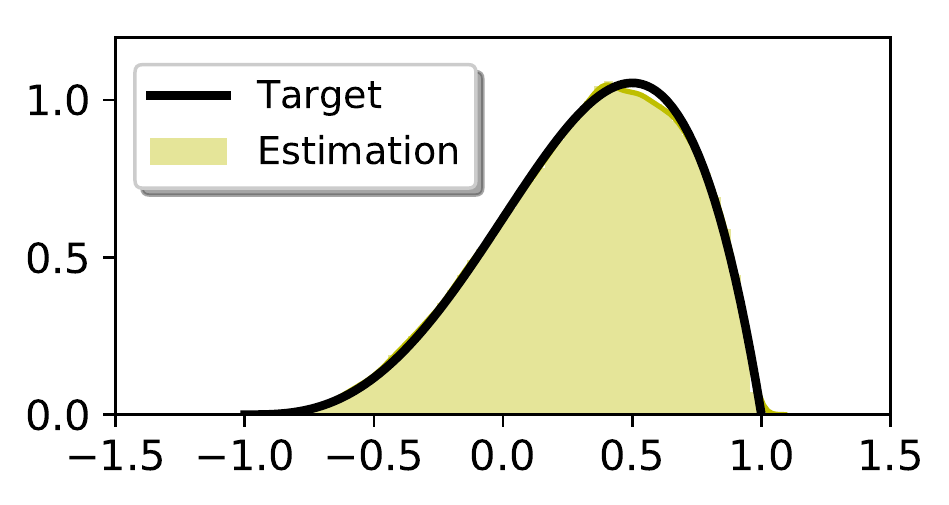}  &
		\hspace{-6mm}
		\includegraphics[width=4.2cm]{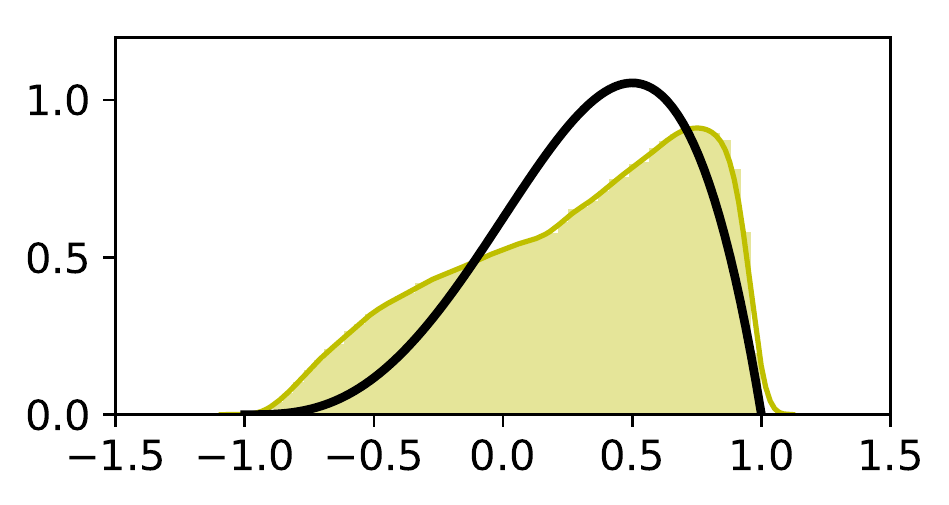}  &
		\hspace{-4mm}
		\includegraphics[width=4.2cm]{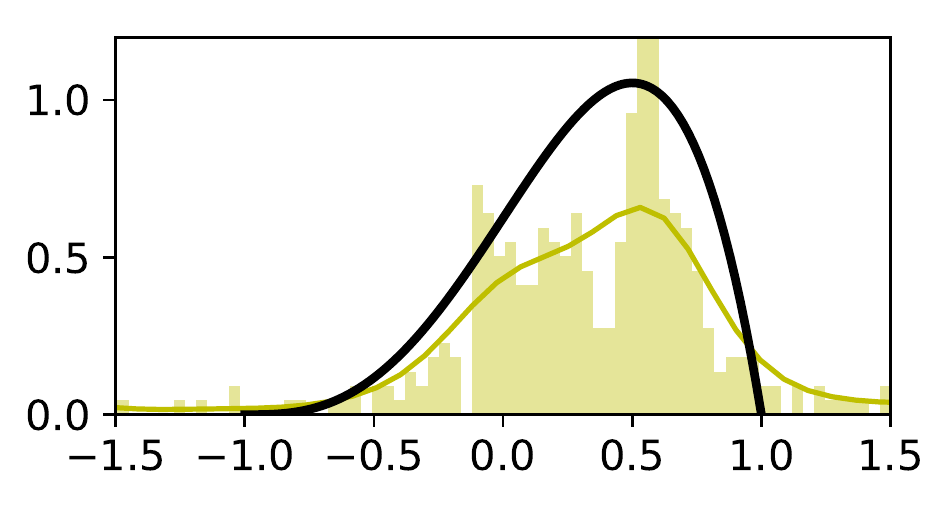}  &
		\hspace{-6mm}
		\includegraphics[width=4.2cm]{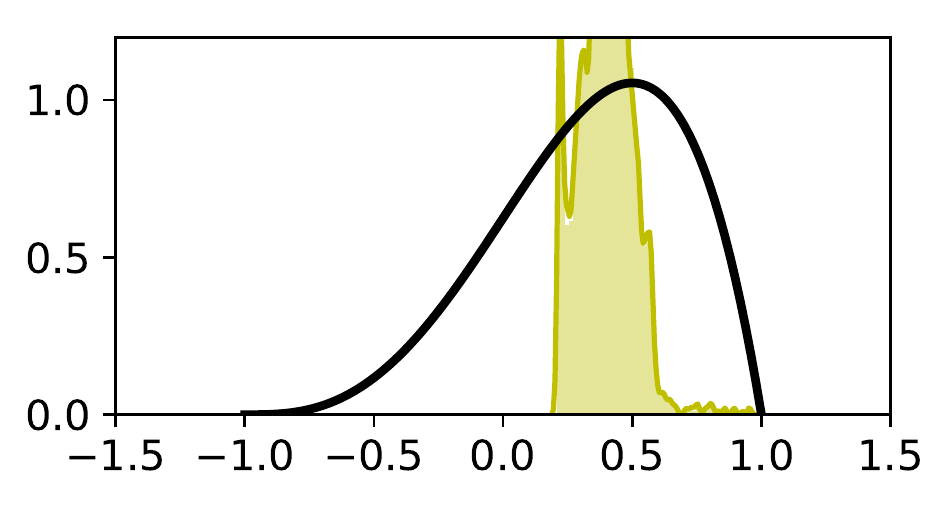} \vspace{-2mm} \\		
		\hspace{-4mm}
		\includegraphics[width=4.2cm]{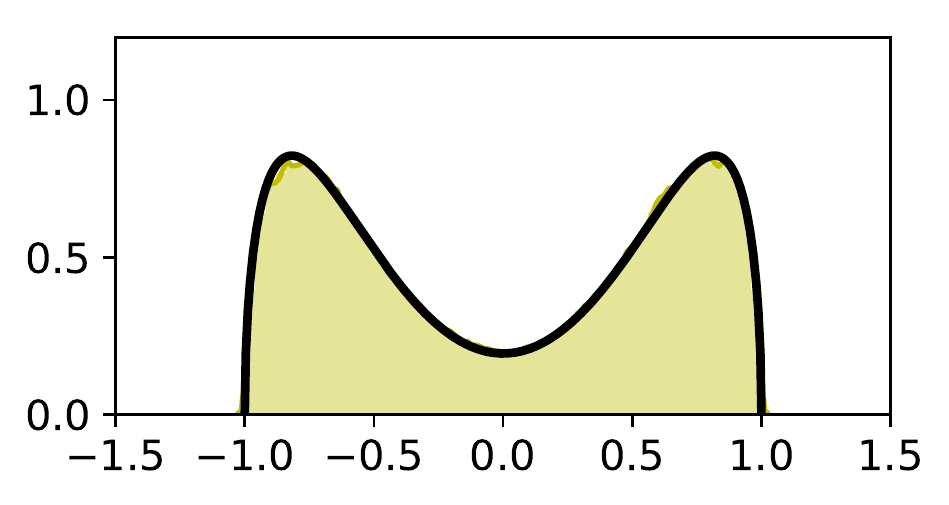}  &
		\hspace{-6mm}
		\includegraphics[width=4.2cm]{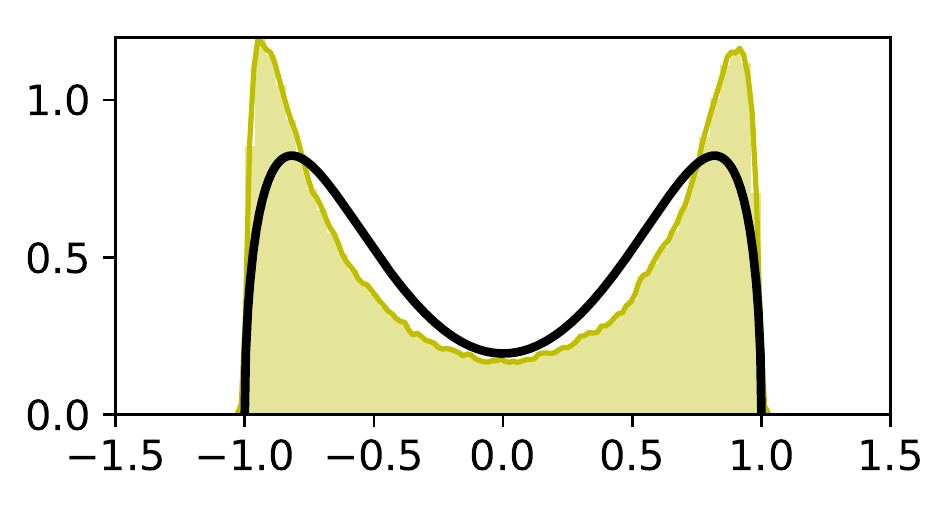}  &
		\hspace{-4mm}
		\includegraphics[width=4.2cm]{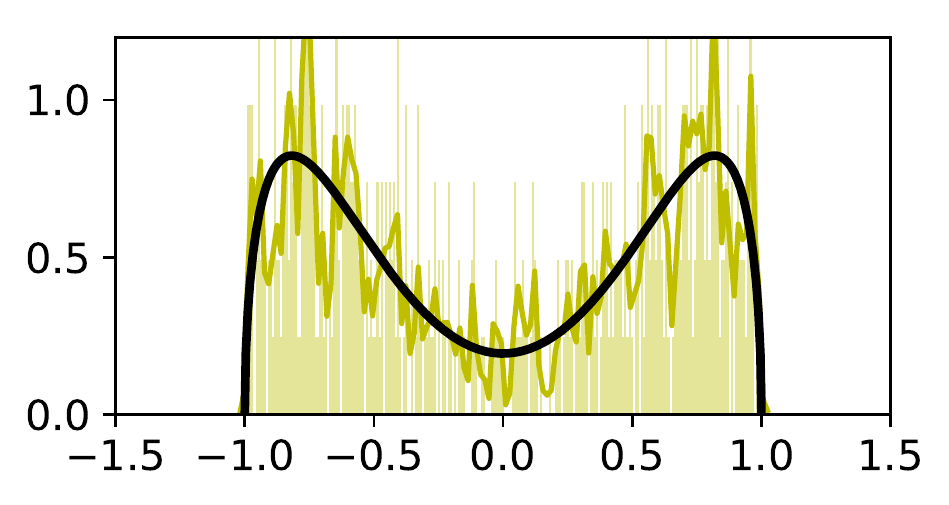}  &
		\hspace{-6mm}
		\includegraphics[width=4.2cm]{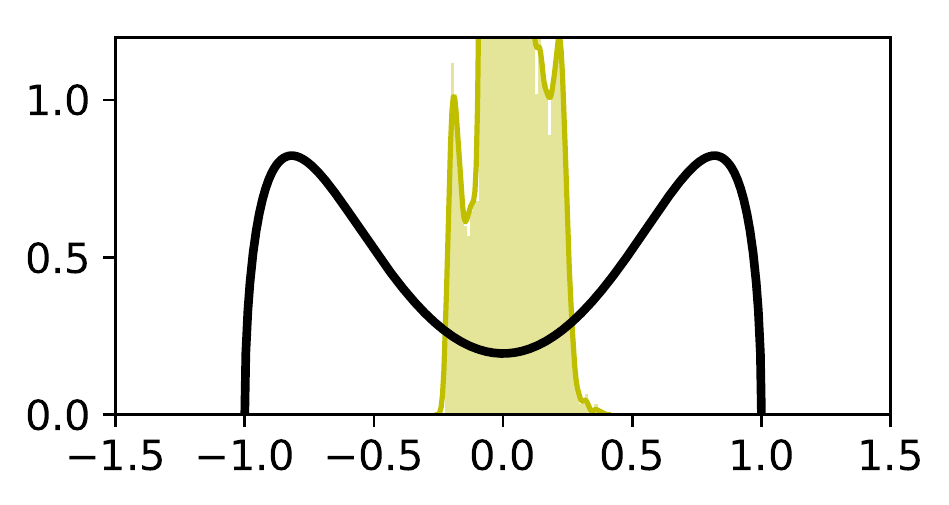}  \\				
		(a) Beta reference \vspace{-0mm} & \hspace{-4mm}
		(b) Gaussian reference \hspace{-0mm} \hspace{-2mm}		&		
		(c) SVGD \vspace{-0mm} & \hspace{-4mm}
		(d) Amortized SVGD \hspace{-0mm} \hspace{-2mm} 
	\end{tabular}
	\vspace{-3mm}
	\caption{\small Sampling from constrained domains}
	\vspace{-4mm}
	\label{fig:constrained}
\end{figure*}

\vspace{-2mm}
\subsection{Sampling from Constrained Domains}
\label{sec:experiments_constrained}
\vspace{-2mm}
To show that RAS can draw samples when $\mathcal{X}$ is bounded, we apply it to sample from the distributions with the support $[-1,1]$. The details for the functions and decay of $\beta$ are in SM. We adopt the Beta distribution as our reference, whose parameters are estimated using the method of moments (see Sec. \ref{sec:learning_distribution}). The activation function in the last layer of the generator is chosen as $\mathtt{tanh}$. As a baseline, we naively use an empirical Gaussian as the reference.
We also compare with the standard SVGD~\cite{Stein} and the amortized SVGD methods \cite{SteinGAN}, in which 512 particles are used. 

Figure~\ref{fig:constrained} shows the comparison. Note that since the support of the Beta distribution is defined in an interval, our RAS can easily match this reference distribution, leading the adversary to accurately estimate the likelihood ratio. Therefore, it closely approximates the target, as shown in Figure~\ref{fig:constrained}(a). Alternatively, when a Gaussian reference is considered, the adversarial ratio estimation can be inaccurate in the low density regions, resulting in degraded sampling performance shown in Figure~\ref{fig:constrained}(b).
Since SVGD is designed for sampling in unconstrained domains, a principled mechanism to extend it for a constrained domain is less clear. Figure~\ref{fig:constrained}(c) shows SVGD results, and a substantial percentage of particles fall out of the desired domain. 
The amortized SVGD method adopts an $\ell_2$ metric to match the generator's samples to the SVGD targets, it collapses to the distribution mode, as in Figure~\ref{fig:constrained}(d).  We observed that the amortized MCMC results~\cite{li2017amcmc,chen2018continuous} are similar to the amortized SVGD~\cite{li2018learning}.

\vspace{-1mm}
\subsection{Soft Q-learning}\label{sec:experiments_sql}
\vspace{-2mm}
Soft Q-learning (SQL) has been proposed recently~\cite{haarnoja2017reinforcement}, with reinforcement learning (RL) policy based on a general class of distributions, with the goal of representing complex, multimodal behavior.
An agent can take an action $a \in\mathcal{A}$ based on a policy $\pi(a|s)$, defined as the probability of taking action $a$ when in state $s$. It is shown in~\cite{haarnoja2017reinforcement} that the target policy has a known unnormalized density $u(a;s)$.

\begin{figure}[t!]
	\vspace{-0mm}
	\begin{tabular}{c c}		
		\hspace{-4mm}
		\includegraphics[width=4.2cm]{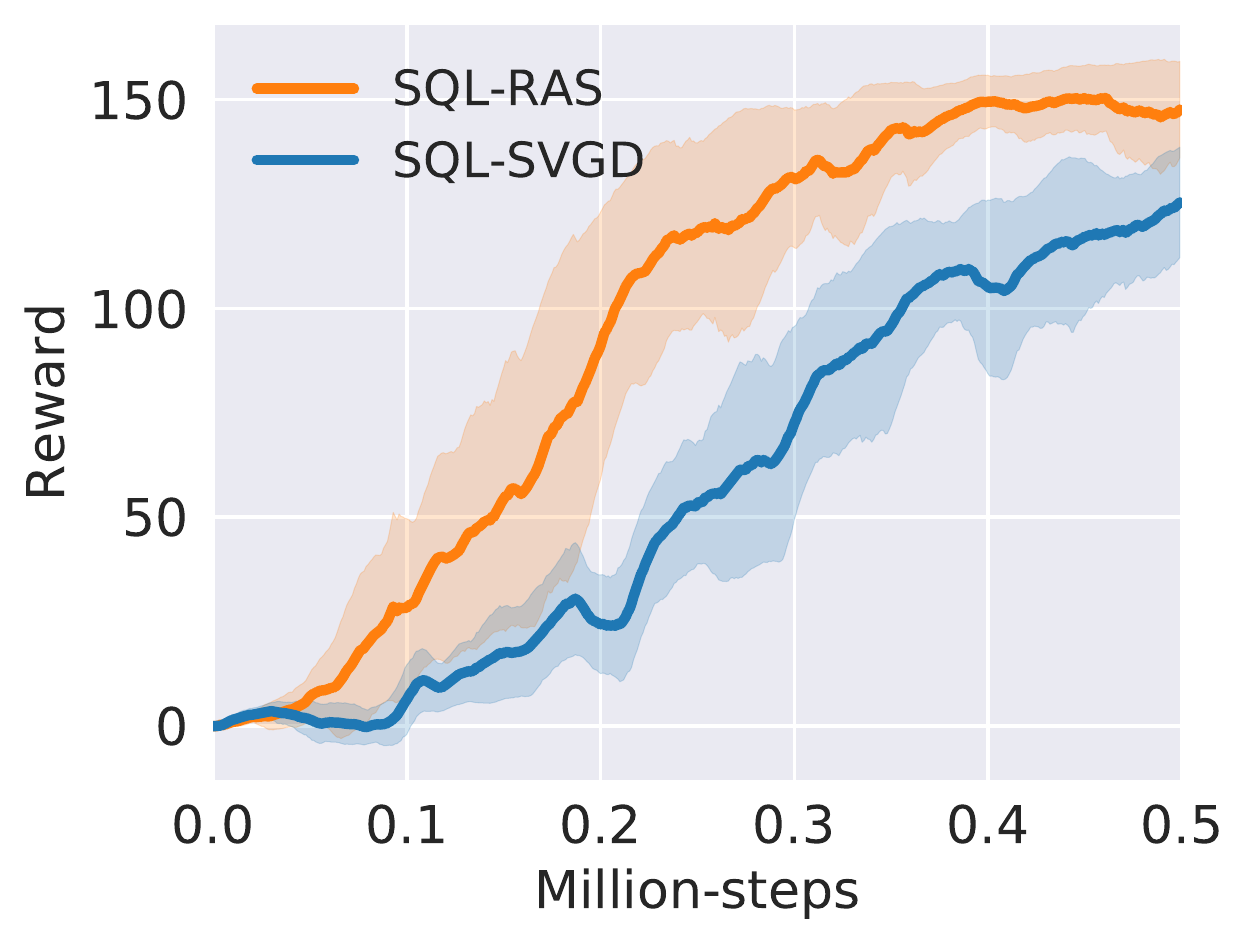}  &
		\hspace{-6mm}
		\includegraphics[width=4.2cm]{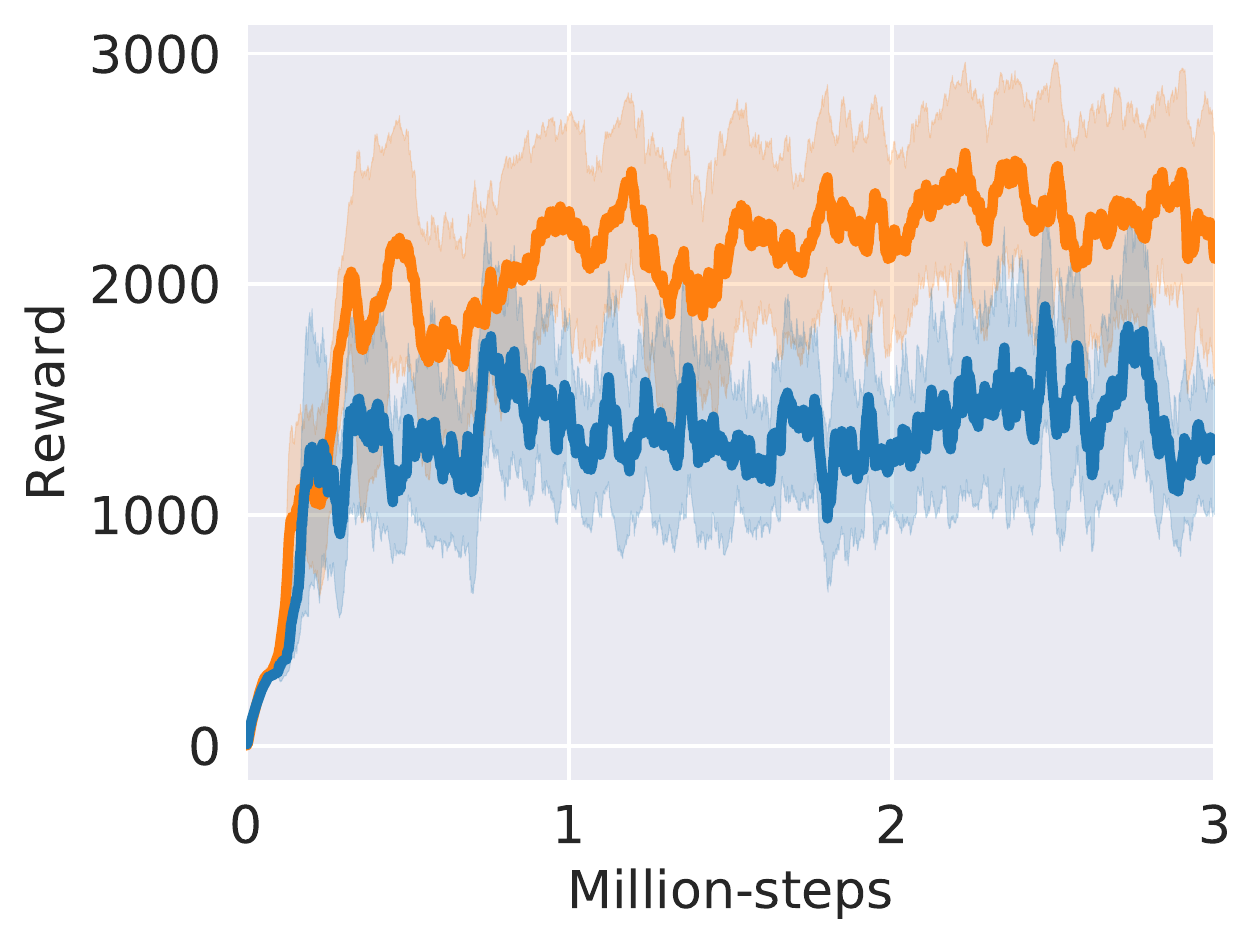}  \vspace{-2mm} \\		
		(a) Swimmer ($\mathtt{rllab}$) \vspace{-0mm} & \hspace{-4mm}
		(b) Hopper-$\mathtt{v1}$ \hspace{-0mm} \hspace{-2mm}\\		
		\hspace{-4mm}
		\includegraphics[width=4.2cm]{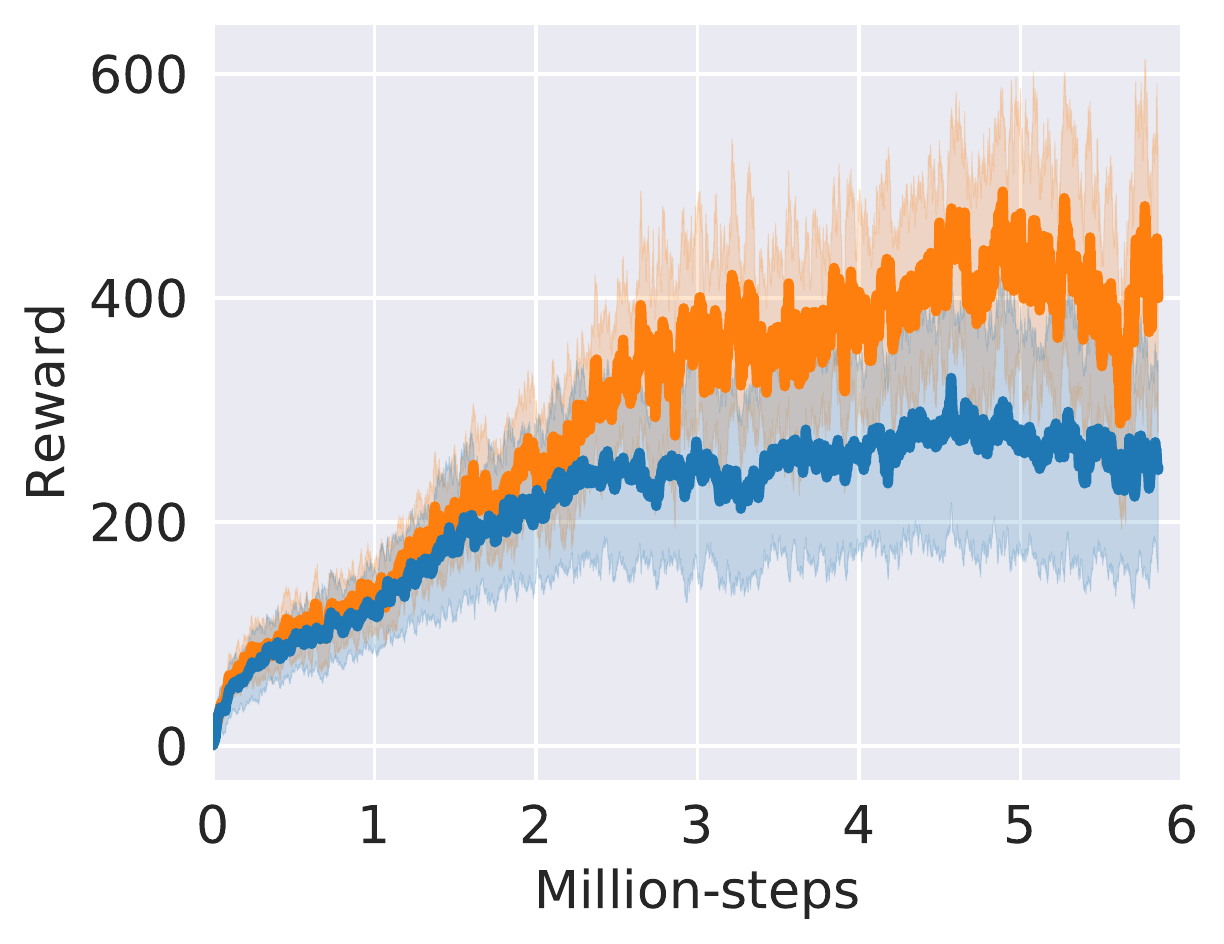}  &
		\hspace{-6mm}
		\includegraphics[width=4.2cm]{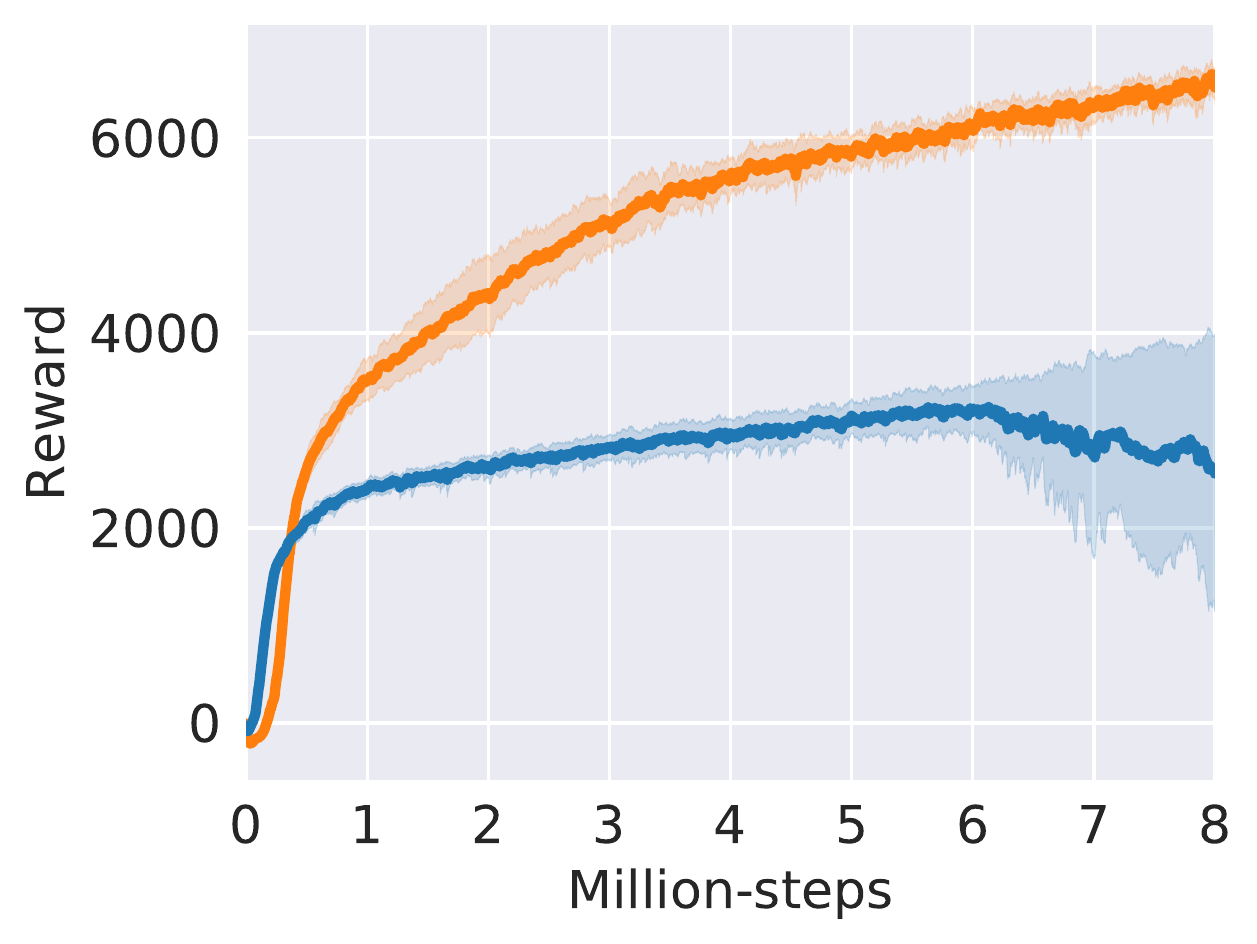}   \vspace{-2mm} \\		
		(c) Humanoid  ($\mathtt{rllab}$) \vspace{-0mm} & \hspace{-4mm}
		(d) Half-cheetah-$\mathtt{v1}$   \hspace{-0mm} \hspace{-2mm} \\
		\hspace{-4mm}
		\includegraphics[width=4.2cm]{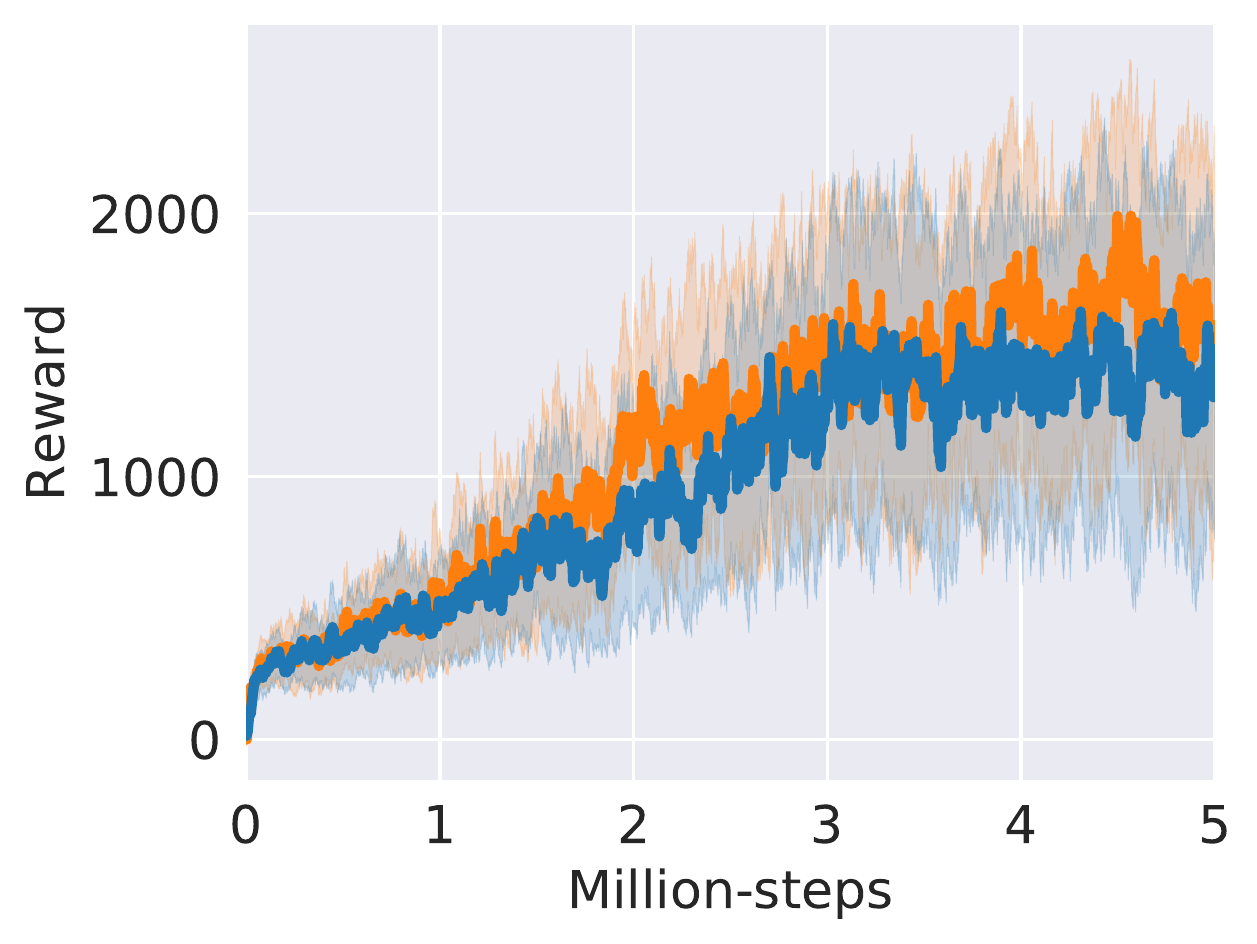}  &
		\hspace{-6mm}
		\includegraphics[width=4.2cm]{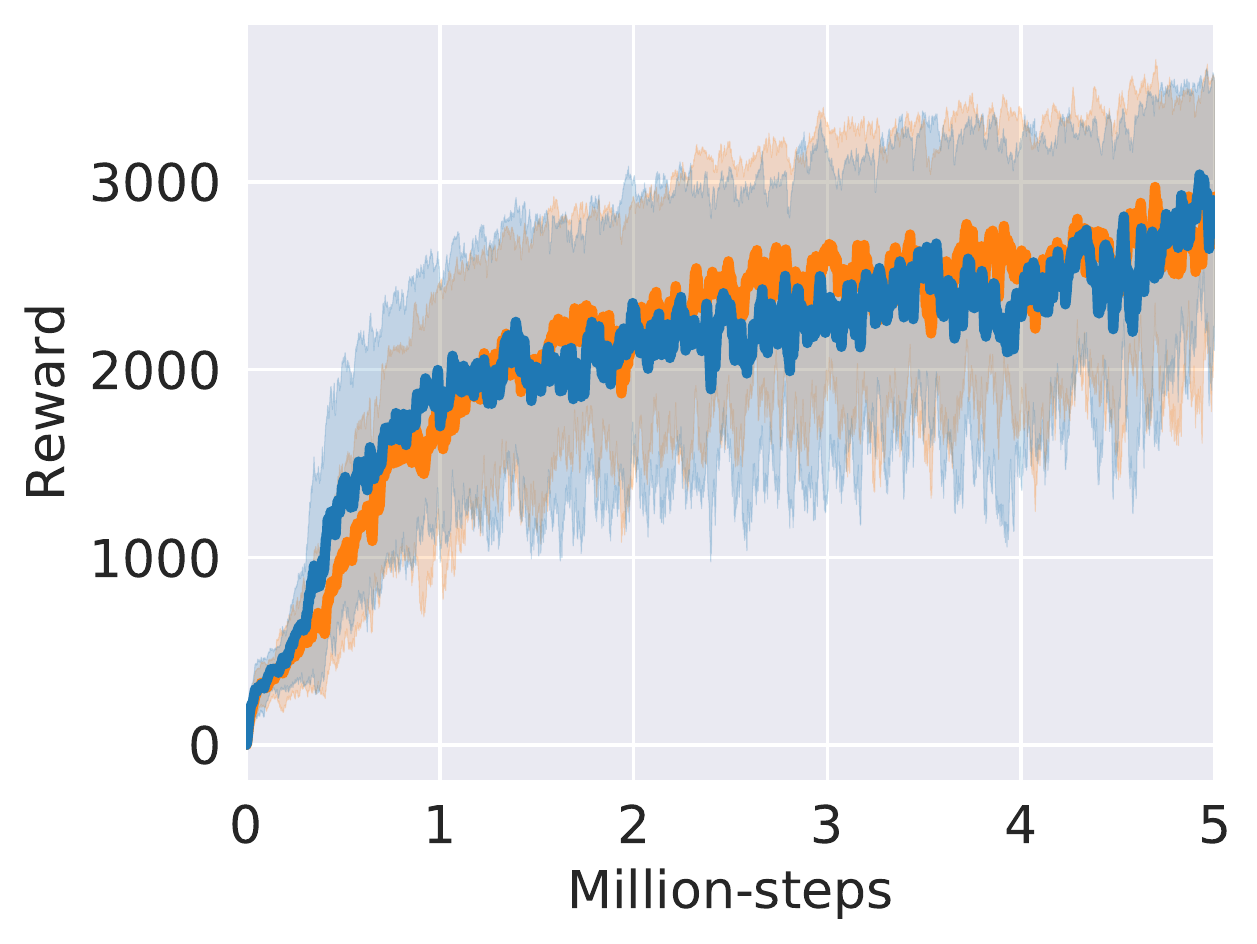}   \vspace{-2mm} \\		
		(e) Ant-$\mathtt{v1}$ \vspace{-0mm} & \hspace{-4mm}
		(f)  Walker-$\mathtt{v1}$  \hspace{-0mm} \hspace{-2mm}
	\end{tabular}
	\vspace{-3mm}
	\caption{\small Soft Q-learning on MuJoCo environments.}
	\vspace{-4mm}
	\label{fig:softQ_MuJoCo}
\end{figure}

\paragraph{SQL-SVGD}  To take actions from the optimal policy (\ie sampling), learning $\theta$ in \cite{haarnoja2017reinforcement} is performed via amortized SVGD in two separated steps: 
$(\RN{1})$ the samples of $u(a|s)$ are first drawn using SVGD by minimizing 
$\mbox{KL}\left(q_{\theta}(a|s)\|u(a;s)\right)$;
$(\RN{2})$ these samples are then used as the target to update $\theta$ under an $\ell_2$ amortization metric. We call this procedure as SQL-SVGD.
\paragraph{SQL-RAS}  Alternatively, we apply our RAS algorithm to replace the amortized SVGD. When the action space is in unconstrained, we may use the Gaussian reference $p_r$. However, the action space is often constrained in continuous control, with each dimension in an interval $[c_1, c_2]$. Hence, we adopt the Beta-distribution reference for RAS.
%

Following~\cite{haarnoja2018soft}, we compare  RAS with amortized SVGD on six continuous
control tasks: Hopper, Half-cheetah, Ant and Walker from the OpenAI $\mathtt{gym}$ benchmark suite~\cite{brockman2016openai}, as well as the Swimmer and Humanoid tasks in the $\mathtt{rllab}$ implementation~\cite{duan2016benchmarking}. Note that the action space is constrained in $[-1,1]$ for all the tasks. The dimension of the action space ranges from 2 to 21 on the different tasks. The higher-dimension environments are usually harder to solve. All hyperparameters used in this experiment are listed in SM.

Figure~\ref{fig:softQ_MuJoCo} shows the total average return of evaluation rollouts during training. We train 3 different instances of each algorithm, with each performing one evaluation rollout every 1k environment steps. The solid curves corresponds to the mean and the shaded regions to the standard derivation
Overall, it show that RAS significantly outperforms amortized SVGD on four tasks both in terms of learning speed and the final performance. This includes the most complex benchmark, the 21-dimensional Humanoid ($\mathtt{rllab}$). On other two tasks, the two methods perform comparably.  In the SQL setting, learning a good stochastic policy with entropy maximization can help training. It means that RAS can better estimate the target policy.

\vspace{-4mm}
\section{Conclusions}
\vspace{-4mm}
We introduce a reference-based adversarial sampling method as a general approach to draw from unnormalized distributions. It allows us to extend GANs from traditional sample-based learning setting to this new setting, and provide novel methods for important downstream applications, \eg Soft Q-learning. RAS can also be easily used for constrained domain sampling.
Further, an entropy regularization is proposed to improve the sample quality, applicable to learning from samples or an unnormalized distribution.  
Extensive experimental results show the effectiveness of the entropy regularization.  
In Soft Q-learning, RAS provides performance comparable to, if not better than, its alternative method amortized SVGD.

\paragraph{Acknowledgements}
We thank Rohith Kuditipudi, Ruiyi Zhang, Yulai Cong and Ricardo Henao for helpful feedback/editing. We acknowledge anonymous reviewers for proofreading and improving the manuscript. The research was supported by DARPA, DOE, NIH, NSF and ONR.

\bibliography{subtex/references,subtex/nips2017}

\begin{thebibliography}{}

\bibitem[Brockman et~al., 2016]{brockman2016openai}
Brockman, G., Cheung, V., Pettersson, L., Schneider, J., Schulman, J., Tang,
  J., and Zaremba, W. (2016).
\newblock Openai gym.
\newblock {\em arXiv preprint arXiv:1606.01540}.

\bibitem[Brooks et~al., 2011]{brooks2011handbook}
Brooks, S., Gelman, A., Jones, G., and Meng, X.-L. (2011).
\newblock {\em Handbook of {M}arkov {C}hain {M}onte {C}arlo}.

\bibitem[Chen et~al., 2018a]{chen2018continuous}
Chen, C., Li, C., Chen, L., Wang, W., Pu, Y., and Duke, L.~C. (2018a).
\newblock Continuous-time flows for efficient inference and density estimation.
\newblock In {\em International Conference on Machine Learning}, pages
  823--832.

\bibitem[Chen et~al., 2018b]{sVAE}
Chen, L., Dai, S., Pu, Y., Li, C., Su, Q., and Carin, L. (2018b).
\newblock Symmetric variational autoencoder and connections to adversarial
  learning.
\newblock {\em AISTATS}.

\bibitem[Chen et~al., 2016]{infogan}
Chen, X., Duan, Y., Houthooft, R., Schulman, J., Sutskever, I., and Abbeel, P.
  (2016).
\newblock Info{GAN}: Interpretable representation learning by information
  maximizing generative adversarial nets.
\newblock In {\em NIPS}.

\bibitem[Duan et~al., 2016]{duan2016benchmarking}
Duan, Y., Chen, X., Houthooft, R., Schulman, J., and Abbeel, P. (2016).
\newblock Benchmarking deep reinforcement learning for continuous control.
\newblock In {\em ICML}.

\bibitem[Feng et~al., 2017]{feng2017learning}
Feng, Y., Wang, D., and Liu, Q. (2017).
\newblock Learning to draw samples with amortized stein variational gradient
  descent.
\newblock {\em UAI}.

\bibitem[Gelman et~al., 1995]{Gelman}
Gelman, A., Carlin, J.~B., S., S.~H., and Rubin, D.~B. (1995).
\newblock Bayesian data analysis.
\newblock {\em London: Chapman and Hall}.

\bibitem[Goodfellow et~al., 2014]{gan}
Goodfellow, I., Pouget-Abadie, J., Mirza, M., Xu, B., Warde-Farley, D., Ozair,
  S., Courville, A., and Bengio, Y. (2014).
\newblock Generative adversarial nets.
\newblock In {\em NIPS}.

\bibitem[Gutmann and Hyv{\"a}rinen, 2010]{noise_contrastive}
Gutmann, M. and Hyv{\"a}rinen, A. (2010).
\newblock Noise-contrastive estimation: A new estimation principle for
  unnormalized statistical models.
\newblock In {\em AISTATS}.

\bibitem[Haarnoja et~al., 2017]{haarnoja2017reinforcement}
Haarnoja, T., Tang, H., Abbeel, P., and Levine, S. (2017).
\newblock Reinforcement learning with deep energy-based policies.
\newblock {\em ICML}.

\bibitem[Haarnoja et~al., 2018]{haarnoja2018soft}
Haarnoja, T., Zhou, A., Abbeel, P., and Levine, S. (2018).
\newblock Soft actor-critic: Off-policy maximum entropy deep reinforcement
  learning with a stochastic actor.
\newblock {\em ICML}.

\bibitem[Hamelryck et~al., 2010]{hamelryck2010potentials}
Hamelryck, T., Borg, M., Paluszewski, M., Paulsen, J., Frellsen, J., Andreetta,
  C., Boomsma, W., Bottaro, S., and Ferkinghoff-Borg, J. (2010).
\newblock Potentials of mean force for protein structure prediction vindicated,
  formalized and generalized.
\newblock {\em PloS one}.

\bibitem[Hastings, 1970]{Hastings}
Hastings, W. (1970).
\newblock Monte {C}arlo sampling methods using {M}arkov {C}hains and their
  applications.
\newblock {\em Biometrika}.

\bibitem[Heusel et~al., 2017]{heusel2017gans}
Heusel, M., Ramsauer, H., Unterthiner, T., Nessler, B., Klambauer, G., and
  Hochreiter, S. (2017).
\newblock {GAN}s trained by a two time-scale update rule converge to a {N}ash
  equilibrium.
\newblock {\em NIPS}.

\bibitem[Hoffman et~al., 2013]{hoffman2013stochastic}
Hoffman, M.~D., Blei, D.~M., Wang, C., and Paisley, J. (2013).
\newblock Stochastic variational inference.
\newblock {\em The Journal of Machine Learning Research}.

\bibitem[Kanamori et~al., 2010]{Kanamori}
Kanamori, T., Suzuki, T., and Sugiyama, M. (2010).
\newblock Theoretical analysis of density ratio estimation.
\newblock {\em IEICE Trans. Fund. Electronics, Comm., CS}.

\bibitem[Kingma and Welling, 2014]{kingma2014vae}
Kingma, D.~P. and Welling, M. (2014).
\newblock Auto-encoding variational {B}ayes.
\newblock {\em ICLR}.

\bibitem[Krizhevsky et~al., 2012]{krizhevsky2012imagenet}
Krizhevsky, A., Sutskever, I., and Hinton, G.~E. (2012).
\newblock Imagenet classification with deep convolutional neural networks.
\newblock In {\em NIPS}.

\bibitem[Li et~al., 2018]{li2018learning}
Li, C., Li, J., Wang, G., and Carin, L. (2018).
\newblock Learning to sample with adversarially learned likelihood-ratio.

\bibitem[Li et~al., 2017a]{ALICE}
Li, C., Liu, H., Chen, C., Pu, Y., Chen, L., Henao, R., and Carin, L. (2017a).
\newblock {ALICE}: Towards understanding adversarial learning for joint
  distribution matching.
\newblock {\em NIPS}.

\bibitem[Li et~al., 2015]{li2015stochastic}
Li, Y., Hern{\'a}ndez-Lobato, J.~M., and Turner, R.~E. (2015).
\newblock Stochastic expectation propagation.
\newblock In {\em NIPS}.

\bibitem[Li et~al., 2017b]{li2017amcmc}
Li, Y., Turner, R.~E., and Liu, Q. (2017b).
\newblock Approximate inference with amortised {MCMC}.
\newblock {\em arXiv preprint arXiv:1702.08343}.

\bibitem[Liu and Wang, 2016]{Stein}
Liu, Q. and Wang, D. (2016).
\newblock Stein variational gradient descent: A general purpose {B}ayesian
  inference algorithm.
\newblock In {\em NIPS}.

\bibitem[Liu et~al., 2015]{liu2015deep}
Liu, Z., Luo, P., Wang, X., and Tang, X. (2015).
\newblock Deep learning face attributes in the wild.
\newblock In {\em ICCV}.

\bibitem[Mescheder et~al., 2017]{AVB}
Mescheder, L., Nowozin, S., and Geiger, A. (2017).
\newblock Adversarial variational {B}ayes: Unifying variational autoencoders
  and generative adversarial networks.
\newblock In {\em ICML}.

\bibitem[Metz et~al., 2017]{metz2017unrolled}
Metz, L., Poole, B., Pfau, D., and Sohl-Dickstein, J. (2017).
\newblock Unrolled generative adversarial networks.
\newblock {\em ICLR}.

\bibitem[Minka, 2001]{minka2001expectation}
Minka, T.~P. (2001).
\newblock Expectation propagation for approximate {B}ayesian inference.
\newblock In {\em UAI}.

\bibitem[Miyato et~al., 2018]{miyato2018spectral}
Miyato, T., Kataoka, T., Koyama, M., and Yoshida, Y. (2018).
\newblock Spectral normalization for generative adversarial networks.
\newblock In {\em ICLR}.

\bibitem[Mohamed and L., 2016]{mohamed2016learning}
Mohamed, S. and L., B. (2016).
\newblock Learning in implicit generative models.
\newblock {\em NIPS workshop on adversarial training}.

\bibitem[Neyman and Pearson, 1933]{neyman1933ix}
Neyman, J. and Pearson, E.~S. (1933).
\newblock On the problem of the most efficient tests of statistical hypotheses.
\newblock {\em Phil. Trans. R. Soc. Lond. A}, 231(694-706):289--337.

\bibitem[Nguyen et~al., 2017]{D2GAN}
Nguyen, T., Le, T., Vu, H., and Phung, D. (2017).
\newblock Dual discriminator generative adversarial nets.
\newblock {\em NIPS}.

\bibitem[Nguyen et~al., 2010a]{likeratio}
Nguyen, X., Wainwright, M., and Jordan, M. (2010a).
\newblock Estimating divergence functionals and the likelihood ratio by convex
  risk minimization.
\newblock {\em IEEE Trans. Info. Theory}.

\bibitem[Nguyen et~al., 2010b]{nguyen2010estimating}
Nguyen, X., Wainwright, M.~J., and Jordan, M.~I. (2010b).
\newblock Estimating divergence functionals and the likelihood ratio by convex
  risk minimization.
\newblock {\em IEEE Transactions on Information Theory}.

\bibitem[Nowozin et~al., 2016]{f-GAN}
Nowozin, S., Cseke, B., and Tomioka, R. (2016).
\newblock f-{GAN}: Training generative neural samplers using variational
  divergence minimization.
\newblock {\em NIPS}.

\bibitem[Oord et~al., 2016]{pixelrnn}
Oord, A., Kalchbrenner, N., and Kavukcuoglu, K. (2016).
\newblock Pixel recurrent neural network.
\newblock In {\em ICML}.

\bibitem[Radford et~al., 2016]{dcgan}
Radford, A., Metz, L., and Chintala, S. (2016).
\newblock Unsupervised representation learning with deep convolutional
  generative adversarial networks.
\newblock In {\em ICLR}.

\bibitem[Rezende et~al., 2014]{rezende2014stochastic}
Rezende, D.~J., Mohamed, S., and Wierstra, D. (2014).
\newblock Stochastic backpropagation and approximate inference in deep
  generative models.
\newblock In {\em ICML}.

\bibitem[Uehara et~al., 2016]{uehara2016generative}
Uehara, M., Sato, I., Suzuki, M., Nakayama, K., and Matsuo, Y. (2016).
\newblock Generative adversarial nets from a density ratio estimation
  perspective.
\newblock {\em arXiv preprint arXiv:1610.02920}.

\bibitem[Van~Trees, 2001]{vanTrees}
Van~Trees, H.~L. (2001).
\newblock {\em Detection, estimation, and modulation theory}.
\newblock John Wiley \& Sons.

\bibitem[Wang and Liu, 2016]{SteinGAN}
Wang, D. and Liu, Q. (2016).
\newblock Learning to draw samples: With application to amortized {MLE} for
  generative adversarial learning.
\newblock In {\em arXiv:1611.01722v2}.

\bibitem[Welling and Teh, 2011]{welling2011bayesian}
Welling, M. and Teh, Y.~W. (2011).
\newblock Bayesian learning via stochastic gradient {L}angevin dynamics.
\newblock In {\em ICML}.

\bibitem[Y.~Pu and Carin, 2017]{Pu_NIPS17}
Y.~Pu, Z.~Gan, R. H. C. L. S.~H. and Carin, L. (2017).
\newblock {VAE} learning via {S}tein variational gradient descent.
\newblock {\em NIPS}.

\bibitem[Zhu et~al., 2017]{cycleGAN}
Zhu, J.-Y., Park, T., Isola, P., and Efros, A. (2017).
\newblock Unpaired image-to-image translation using cycle-consistent
  adversarial networks.
\newblock {\em ICCV}.

\end{thebibliography}
\bibliographystyle{plainnat}

\newpage
\twocolumn[
\aistatstitle{\Large Supplementary Material : Adversarial Learning of a Sampler\\Based on an Unnormalized Distribution}
]

\appendix

\section{Proof of the Entropy Bound in Lemma 1}
Consider random variables $(x,\epsilon)$ under the joint distribution $q_{\theta}(x,\epsilon)=q(\epsilon)q_{\theta}(x|\epsilon)$, where $q_{\epsilon}(x|\epsilon)=\delta(x-h_{\theta}(\epsilon))$. The mutual information between $x$ and $\epsilon$ satisfies $ I(x;\epsilon)= H(x)- H(x|\epsilon)= H(\epsilon)- H(\epsilon|x)$. 
Since $q_{\theta}(x|\epsilon)$ is a deterministic function of $\epsilon$, $ H(x|\epsilon)=0$. We therefore have $H(x)= H(\epsilon)- H(\epsilon|x)$, where $ H(\epsilon)=-\int q(\epsilon)\log q(\epsilon) \rm{d} \epsilon$ is a constant wrt $\theta$. For general distribution $t_{\xi}(\epsilon|x)$,
\begin{align}
& H(\epsilon|x)=-\mathbb{E}_{p_{\theta}(x,\epsilon)}\log p_{\theta}(\epsilon|x)\\
=& -\mathbb{E}_{q_{\theta}(x,\epsilon)}\log t_{\xi}(\epsilon|x)-\mathbb{E}_{q_{\theta}(x)}  \mbox{KL}(q_{\theta} (\epsilon|x) \| t_{\xi}(\epsilon|x)) \nonumber\\
\leq & -\E_{q_{\theta}(x,\epsilon)}\log t_{\xi}(\epsilon|x)
\end{align}
We consequently have
\begin{align}
H(x) & =-\mathbb{E}_{q_{\epsilon}(x)}\log q_{\theta} (x){\rm d}x \nonumber \\
& =H(\epsilon)- H(\epsilon|x)\geq H(\epsilon)+\mathbb{E}_{p_{\theta} (x,\epsilon)}\log t_{\xi} (\epsilon|x).
\end{align}
Therefore, entropy is lower bounded by the log likelihood or negative cycle-consistency loss; minimizing the cycle-consistency loss maximizes the entropy or mutual information.
$\square$

\section{Experiments}

\subsection{Sampling from 8-GMM\label{sm:8gmm}}

Two methods are presented for estimating the likelihood ratio: (i) ${\sigma}$-ALL for the discriminator in the standard GAN \ie Eq~\eqref{eq:phi}; (ii) $f$-ALL for a variational characterization of $f$-measures in~\cite{likeratio}. 

\begin{figure*}[t!] \centering
	\vspace{-0mm}
	\begin{tabular}{c }
		\includegraphics[width=15.0cm]{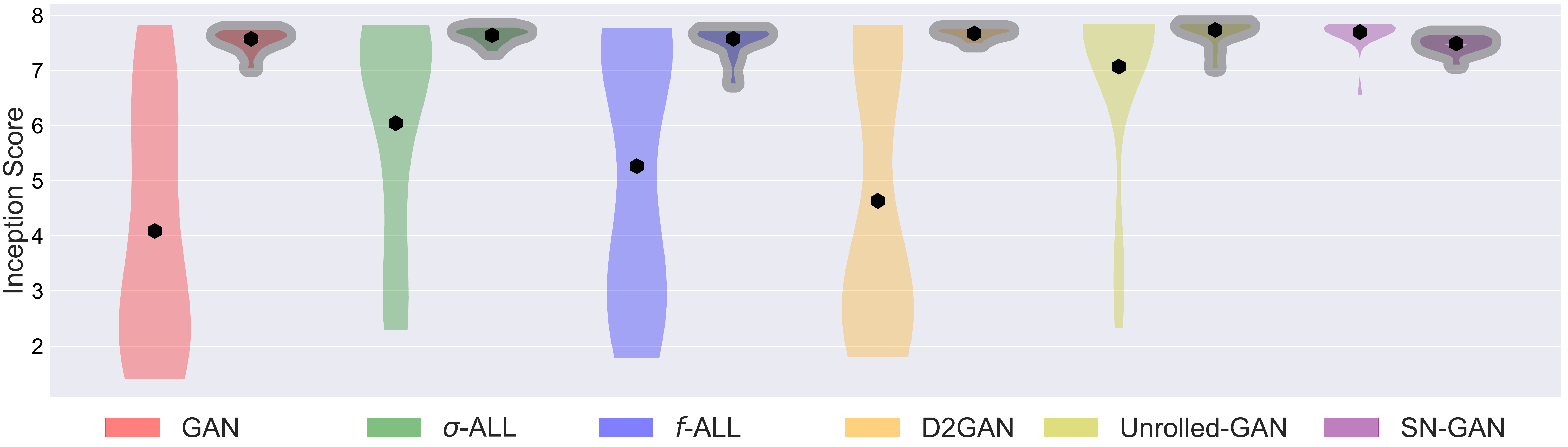} 
	\end{tabular}
	\vspace{-3mm}
	\caption{Comparison of inception score on different GAN variants. The GAN variants and their corresponding entropy-regularized variants  are visualized in the same color, with the latter shaded slightly. The balck dots indicate the means of the distributions.}
	\vspace{-0mm}
	\label{fig:8gmm_comparison_icp}
\end{figure*}

In Figure~\ref{fig:8gmm_comparison_icp}, we plot the distribution of inception score (ICP) values~\cite{ALICE}. Similar conclusions as in the case of the symmetric KL divergence metric the  can be drawn: (1) The likelihood ratio impelmentation improve the original GAN, and (2) the entropy regularizer improve the all GAN variants. Note that because ICP favors the samples closer to the mean of each mode and SN-GAN generate samples that concentrate only around the mode’s centroid, SN-GAN show slightly better ICP than its entroy-regularized version. We argue that the entropy regualizer help gernerate diverse samples, the lower value of  ICP is just due to the limitation of the metric. 

\begin{figure*}[t!] \centering
	\vspace{-0mm}
	\begin{tabular}{cc }
		\hspace{2mm}
		\begin{minipage}{8.0cm}\vspace{0mm}
			\includegraphics[width=7.0cm]{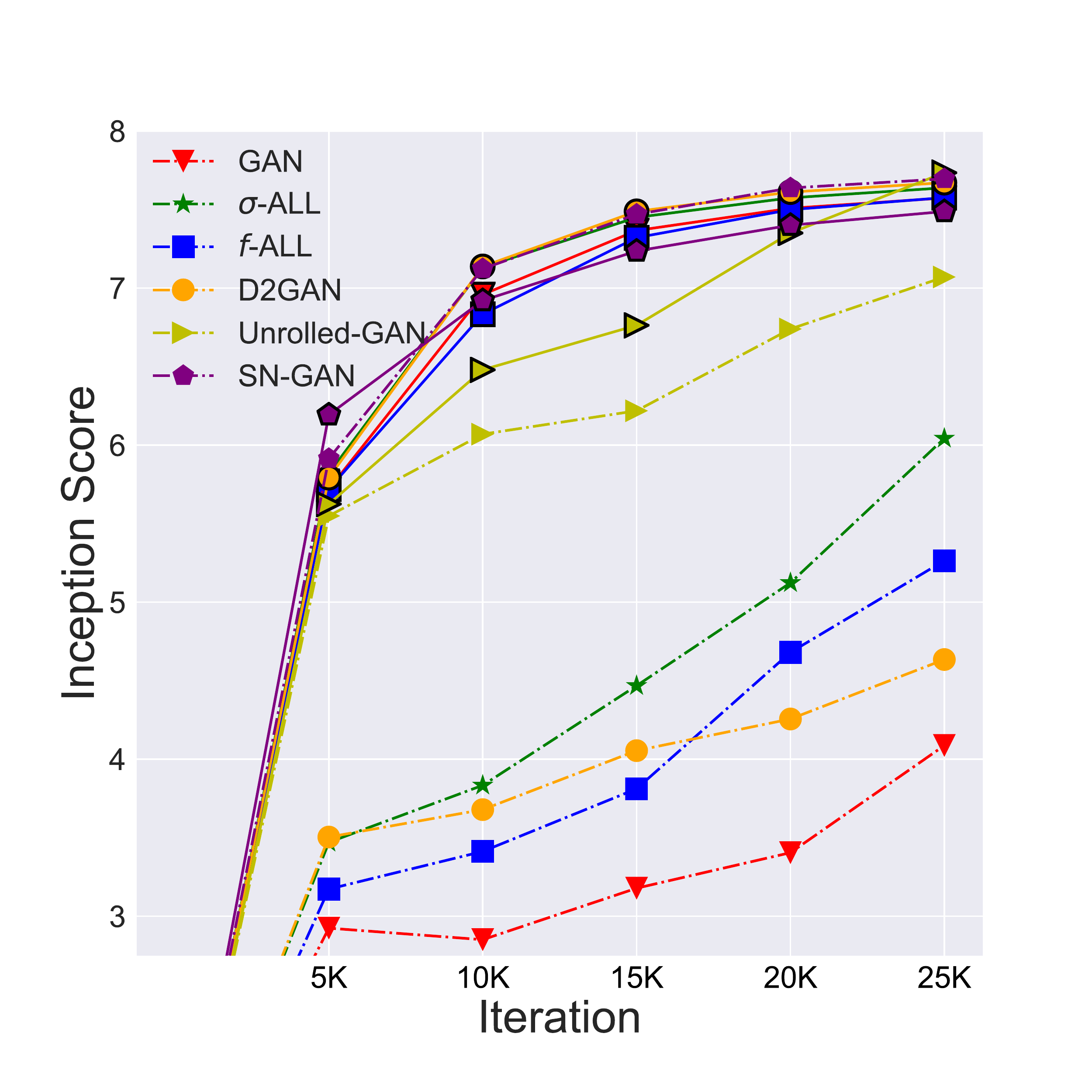} 
		\end{minipage}  & 
		\begin{minipage}{8.0cm}\vspace{0mm}
			\includegraphics[width=7.0cm]{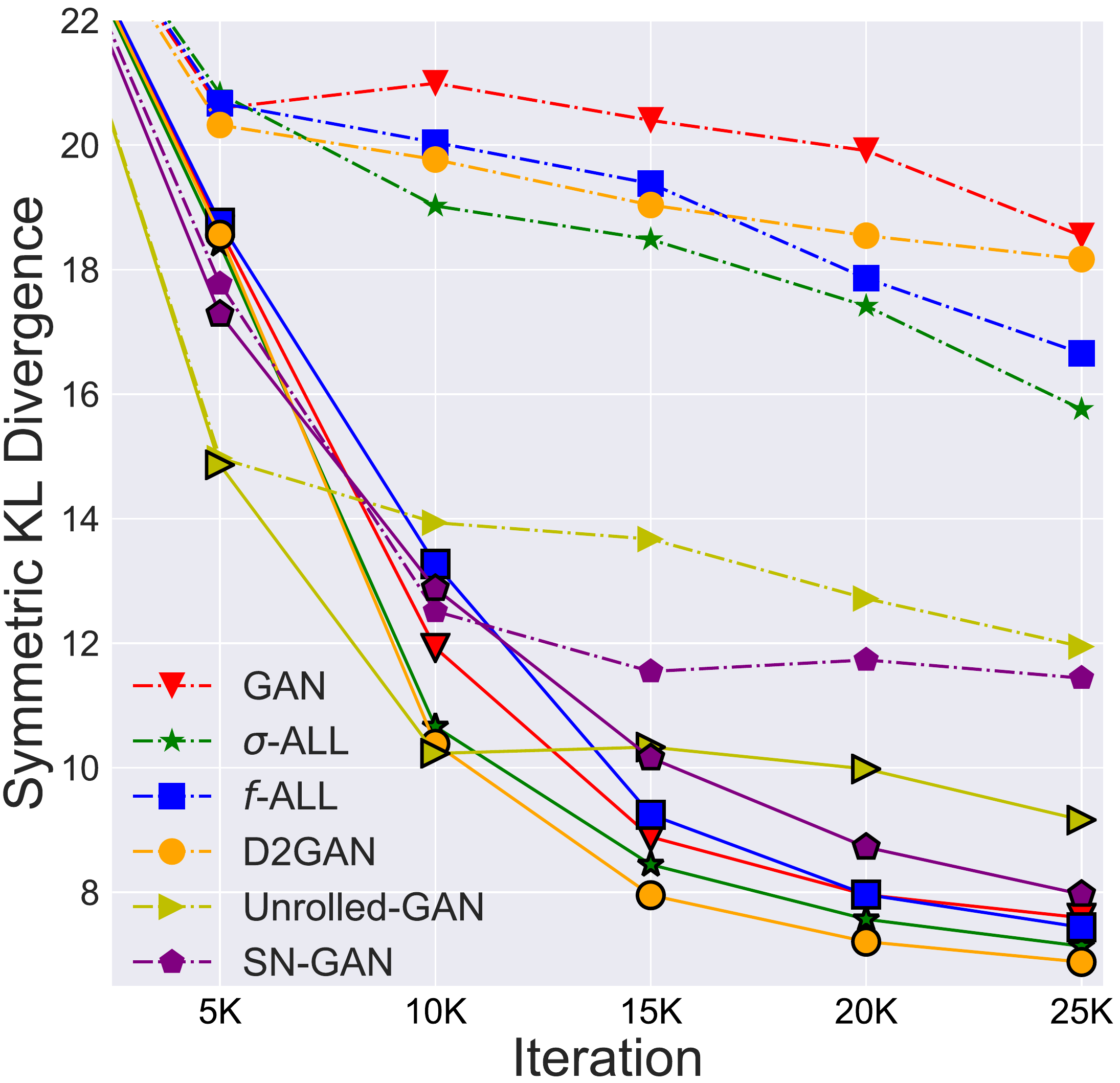} 
		\end{minipage} \\ 
		(a) Inception score over iterations.
		& 
		(b) Symmetric KL over iterations.  \\
	\end{tabular}
	\vspace{-0mm}
	\caption{Learning curves of different GAN variants. The standard GAN variants are visualized as dashed lines, while their corresponding entropy-regularized variants are visualized as the solid lines in the same color.}
	\vspace{-0mm}
	\label{fig:8gmm_learning_curves}
\end{figure*}

The learning curves of the inception score and symmetric KL divergence values are plot over iterations in Figure~\ref{fig:8gmm_learning_curves} (a) and (b), respectively. The family of GAN variants with entropy term dominate the performance, compared with those without the entropy term. We conclude that the entropy regularizer can significantly improve the convergence speed and the final performance.

\paragraph{Architectures and Hyper-parameters\label{sm:networks}}
For the 8-GMM and MNIST datasets, the network architectures are specified in Table~\ref{tab:archit}, and hyper-parameters are detailed in Table~\ref{tab:hyperparas}. The inference network is used to construct the cycle-consistency loss to bound the entropy.

\begin{table}[h!]\centering
	\vspace{3mm}
	\begin{minipage}{8.3cm}\vspace{0mm}	\centering
		\caption{ The convention for the architecture ``X--H--H--Y'': X is the input size, Y is the output size, and H is the hidden size. ``ReLU'' is used for all hidden layer, and the activation of the output layer is linear, except the generator on MNIST is the sigmoid }
		\vspace{-1mm}
		\label{tab:archit}
		\vskip 0.0in 
		\centering
		\small
		\hspace{ 0mm} 	
		\begin{adjustbox}{scale=1.0,tabular=l|c|c,center}
			&  8-GMM &   MNIST    \\  \hline
			Networks  &   \hspace{-2mm} Size \hspace{-3mm} 
			& \hspace{-2mm} Size \hspace{-3mm}\\
			\hline
			Generator	
			& $2$--$128$--$128$--$2$	
			& $32$--$256$--$256$--$784$	\\
			Discriminator 		
			& $2$--$128$--$128$--$1$	  
			& $784$--$256$--$256$--$1$  \\
			Auxiliary 		
			& $2$--$128$--$128$--$2$	   
			&  $784$--$256$--$256$--$32$   
		\end{adjustbox}
	\end{minipage}
	\vspace{3mm}
\end{table}
\begin{table}[h!]\centering
	\vspace{3mm}
	\begin{minipage}{8.3cm}\vspace{0mm}	\centering
		\caption{ The hyper-parameters of experiments. Adam optimizer is used.}
		\vspace{-1mm}
		\label{tab:hyperparas}
		\vskip 0.0in 
		\centering
		\small
		\hspace{ 0mm} 	
		\begin{adjustbox}{scale=0.95,tabular=l|cc,center}
			Hyper-parameters & 8GMM  &  MNIST   \\ 
			\hline
			Learning rate & $2 \! \times \!10^{-4}$  & $1 \! \times \!10^{-3}$    \\
			Batch Size & $ 1024 $   & $ 64 $ \\
			$\#$Updates & $50$k Iterations   & $60$ Epoches 
		\end{adjustbox}
	\end{minipage}
	\vspace{3mm}
\end{table}

We further study three real-world datasets of increasing diversity and size: MNIST, CIFAR10~\cite{krizhevsky2012imagenet} and CelebA~\cite{liu2015deep}. For each dataset, we start with a standard GAN model: two-layer fully connected (FC) networks on MNIST, as well as DCGAN~\cite{dcgan} on CIFAR and CelebA. We then add the entropy regularizer. 
On MNIST, we repeat the experiments 5 times, and the mean ICP is shown.
On CIFAR and CelebA, the performance is also quantified via the recently proposed Fr\'{e}chet Inception Distance (FID)~\cite{heusel2017gans}, which approximates the Wasserstein-2 distance of generated samples and true samples.
The best ICP and FID for each algorithm are reported in Table~\ref{tab:allen_icp}. 
The entropy variants consistently show better performance than the original counterparts.

\begin{table}[h!]\centering
	\vspace{-0mm}
	\begin{minipage}{8.3cm}\vspace{0mm}	\centering
		\caption{\small Performance of entropy regularization. Results marked with [${\star}$] and [${\diamond}$] are from~\cite{D2GAN} and ~\cite{heusel2017gans}, respectively. }
		\vspace{-1mm}
		\label{tab:allen_icp}
		\vskip 0.0in 
		\centering
		\small
		\hspace{ 0mm} 	
		\begin{adjustbox}{scale=1.0,tabular=l|cc|cc,center}
			& \multicolumn{2}{c|}{ICP $\uparrow$} 	& \multicolumn{2}{c}{FID  $\downarrow$}    \\  \hline
			Dataset  &   \hspace{-2mm} Standard \hspace{-3mm} &  E$_{cc}$ 
			& \hspace{-2mm} Standard \hspace{-3mm} &  E$_{cc}$  \\
			\hline
			MNIST	&$7.24~~$ 		& ${\bf 8.08}$          			   & - & - \\
			CIFAR 		&$6.40^{\star}$		& ${\bf 6.86}$            			   & $36.90^\diamond$ & ${\bf 36.70}$\\
			CelebA 		& -	& -           			   & $12.50^\diamond$ & ${\bf 11.88}$
		\end{adjustbox}
	\end{minipage}
	\vspace{-0mm}
\end{table}

\vspace{-5mm}
\subsection{Constrained Domains\label{sm:constrained}}
\vspace{-4mm}
%
The two functions are:
(1) $u_1(x) = \max( (1- (x/2 + 0.5) )(x/2 + 0.5)^3, 0 )$, and 
(2) $u_2(x) = \max( (1- (x/2 + 0.5) )^{0.5}  (x/2 + 0.5)^5 + (1- (x/2 + 0.5) )^5  (x/2 + 0.5)^{0.5} , 0 )$. 
The network architectures used for constrained domains are reported in Table~\ref{tab:archit_constrained}.  The batch size is 512, learning rate is $1\times10^{-4}$. The total training iterations $T=20$k, and we start to decay $\beta$ after $T_0=10$k iterations.
\begin{table}[h!]\centering
	\vspace{-0mm}
	\begin{minipage}{8.3cm}\vspace{0mm}	\centering
		\caption{ The convention for the architecture ``X--H--H--Y'': X is the input size, Y is the output size, and H is the hidden size. ``ReLU'' is used for all hidden layer, and the activation of the output layer is ``Tanh''. }
		\vspace{-1mm}
		\label{tab:archit_constrained}
		\vskip 0.0in 
		\centering
		\small
		\hspace{ 0mm} 	
		\begin{adjustbox}{scale=1.0,tabular=l|c,center}
			Networks  &   \hspace{-2mm} Size \hspace{-3mm}  \hspace{-3mm}\\
			\hline
			Generator	
			& $2$--$128$--$128$--$1$	\\
			Discriminator 		
			& $2$--$128$--$128$--$1$	    \\
			Auxiliary 		
			& $1$--$128$--$128$--$2$	  
		\end{adjustbox}
	\end{minipage}
	\vspace{-3mm}
\end{table}

\begin{algorithm*}[htb]
	\caption{Adversarial Soft Q-learning}
	\label{alg:A_SQL}
	\begin{algorithmic}[1]
		\REQUIRE  {Create replay memory $\mathcal{D} = \emptyset$; Initialize network parameters $ \thetav, \phiv, \psiv$; Assign target parameters: $\overline{\thetav} \leftarrow \thetav$, $\overline{\psiv} \leftarrow \psiv$.}
		\FOR{each epoch}
		\FOR{each $t$}
		\vspace{2mm}
		\STATE  $\color{blue}\mathtt{\%~Collect~expereince}$
		\STATE  {Sample an action for $\sv_{t}$ using $g^{\thetav}$: $\av_{t} \leftarrow g^{\thetav}(\xiv; \sv_{t})$, where $\xiv \sim \mathcal{N}(\mathbf{0}, \Imat)$.}
		\STATE {Sample next state and reward from the environment: 
			$\sv_{t+1} \sim P_{\sv}$ and $r_{t} \sim P_r$ 
		}
		\STATE  {Save the new experience in the replay memory: $\mathcal{D} \leftarrow  \mathcal{D}\cup \{\sv_{t}, \av_t, r_t, \sv_{t+1}\}$ }
		\vspace{2mm}
		\STATE $\color{blue}\mathtt{\%~Sample~a~minibatch~from~the~replay~memory}$
		\STATE {$\{(\sv_{t}^{(i)}, \av_t^{(i)}, r_t^{(i)}, \sv_{t+1}^{(i)})\}^{n}_{i=0} \sim \mathcal{D}$.}
		\vspace{2mm}
		\STATE $\color{blue}\mathtt{\%~Update~Q~value~network}$
		\STATE  {Sample $\{\av^{(i,j)}\}_{j=0}^{M} \sim q_{\av '}$ for each $\sv^{(i)}_{t+1}$.}
		\STATE {Compute the soft Q-values $u(\av, \sv)$ as the target unnormalized density form.
		}
		\STATE {Compute gradient of Q-network and update $\psiv$ \\
	}
		\vspace{2mm}
		\STATE  $\color{blue}\mathtt{\%~Update~policy~network~via~RAS}$
		\STATE  {Sample actions for each $\sv_t^{(i)}$ from the stochastic policy via  \\
			\hspace{30mm}
			$\av_{t}^{(i,j)} = f^{\phiv}(\xiv^{(i,j)}, \sv_t^{(i)})$, where $\{\xiv^{(i,j)}\}^M_{j=0} \sim \mathcal{N}(\mathbf{0}, \Imat)$}
		
		\STATE  {Sample actions for each $\sv_t^{(i)}$ from a Beta (or Gaussian) reference policy $\{\av_r^{(i,j)}\}^M_{j=0}  \sim p_r(\av| \sv_t^{(i)} )$ }		
		\STATE {Compute gradient  of discriminator in \eqref{eq:phi_ras} and update $\phiv$}
		
		\STATE  {Compute gradient of policy network in \eqref{eq:theta_ras}, and update $\thetav$}
		\ENDFOR
		\IF {epoch {\it mod} update\_interval = 0}
		\STATE{Update target parameters: $\overline{\thetav} \leftarrow \thetav$, $\overline{\psiv} \leftarrow \psiv$}	
		\ENDIF
		\ENDFOR
	\end{algorithmic}
\end{algorithm*}

\subsection{Soft Q-learning\label{sm:sql}}
We show the detailed setting of environments in Soft Q-Learning in Table~\ref{tab:env_params}. 
The network architectures are specified in Table~\ref{tab:archit_softQ}, and hyper-parameters are detailed in Table~\ref{tab:hyperparas_softQ}. We only add the entropy regularization at the beginning to stabilize training, and then quickly decay $\beta$ to 0. The total training epoch is 200, and we start to decay $\beta$ after 10 epochs, and set it $0$ after 50 epochs. This is because we observed that the entropy regularization did not help in the end, and removing it could accelerate training.

\begin{table}[h!]
	\renewcommand{\arraystretch}{0.8}
	\centering
	\caption{Hyper-parameters in SQL.}
	\label{tab:env_params}
	\vspace{1mm}
	\begin{tabular}{ r|ccc }
		\hline
		Environment   &Action &Reward   & Replay  \\ 
		&Spcae &Scale  & Pool Size\\
		\hline
		Swimmer (rllab)  &2   & 100          & $10^6$\\
		Hopper-v1    &3  & 1            & $10^6$\\
		HalfCheetah-v1   &6  & 1          & $10^7$\\
		Walker2d-v1   &6   & 3            & $10^6$\\
		Ant-v1     &8  & 10        & $10^6$\\
		Humanoid (rllab)  &21  & 100         & $10^6$\\
		\hline
	\end{tabular}
\end{table}

\begin{table}[h!]\centering
	\vspace{2mm}
	\begin{minipage}{8.3cm}\vspace{0mm}	\centering
		\caption{ The convention for the architecture ``X--H--H--Y'': X is the input size, Y is the output size, and H is the hidden size. ``ReLU'' is used for all hidden layer, and the activation of the output layer is ``Tanh" for the policy network and linear for the others. $\Scal$ represents the state, $\mathcal{A}$ represents the action. $\Ncal$ is the gaussian noise.  The dimension of the noise is the same as the action space.  The parameters settings of SVGD version and ours version are the same.}
		\vspace{-1mm}
		\label{tab:archit_softQ}
		\vskip 0.0in 
		\centering
		\small
		\hspace{ 0mm} 	
		\begin{adjustbox}{scale=1.0,tabular=l|l,center}
			Networks  &   \hspace{-2mm} Size \hspace{-3mm}  \\
			\hline
			Policy-Network
			& $|\mathcal{S+N}|$--$128$--$128$--$|\mathcal{A}|$	 \\
			\hline	
			Q-Network 		
			& $|\mathcal{S+A}|$--$128$--$128$--$1$	\\   
			\hline	
			Inverse Mapping
			& $|\mathcal{A}|$--$128$--$128$--$|\mathcal{S}+\mathcal{N}|$	\\  
			\hline	
			Discriminator 		
			& $|\mathcal{A+S}|$--$128$--$128$--$128$--$1$	
		\end{adjustbox}
	\end{minipage}
	\vspace{-0mm}
\end{table}

\begin{table}[h!]\centering
	\vspace{1mm}
	\begin{minipage}{8.3cm}\vspace{0mm}	\centering
		\caption{ The hyper-parameters of experiments.}
		\vspace{-1mm}
		\label{tab:hyperparas_softQ}
		\vskip 0.0in 
		\centering
		\small
		\hspace{ 0mm} 	
		\begin{adjustbox}{scale=0.95,tabular=l|cc,center}
			Hyper-parameters & Values\\
			\hline
			Learning rate of Policy & $3 \! \times \!10^{-4}$ \\
			Learning rate of Q-network & $3 \! \times \!10^{-4}$  \\
			Batch Size & $ 128 $\\
			$\#$Particle in SVGD & $32$
		\end{adjustbox}
	\end{minipage}
	\vspace{-0mm}
\end{table}

\end{document}